\documentclass{article}

\PassOptionsToPackage{numbers, sort&compress}{natbib}
 \usepackage[preprint]{neurips_2026}

\usepackage[utf8]{inputenc} 
\usepackage[T1]{fontenc}    
\usepackage{hyperref}       
\usepackage{url}            
\usepackage{booktabs}       
\usepackage{amsfonts}       
\usepackage{nicefrac}       
\usepackage{microtype}      
\usepackage{xcolor}         

\usepackage{amsmath}
\usepackage{amssymb}
\usepackage{mathtools}
\usepackage{amsthm}
\usepackage[capitalize]{cleveref}
\usepackage{url}
\usepackage{makecell}
\usepackage{xspace}
\usepackage{dsfont}
\usepackage{afterpage}

\usepackage{enumitem}
\usepackage{booktabs}
\usepackage{caption} 
\usepackage{multirow} 
\usepackage{enumitem}
\usepackage{colortbl} 

\usepackage{wrapfig}
\usepackage{duckuments}

\usepackage{subcaption} 
\usepackage{tikz}
\usepackage{mwe}

\usepackage{listings}
\usepackage{graphicx,color}
\usepackage{listings}
\usepackage{xcolor}

\definecolor{cornellred}{rgb}{0.7, 0.11, 0.11}
\definecolor{cadmiumgreen}{rgb}{0.0, 0.42, 0.24}
\definecolor{aliceblue}{rgb}{0.91, 0.94, 0.97}
\definecolor{darkblue}{rgb}{0.83, 0.89, 0.97}
\definecolor{Red}{rgb}{0.941, 0.243, 0.243}
\definecolor{Green}{RGB}{55, 178, 77}
\definecolor{Blue}{rgb}{0.098,0.3,0.9}
\definecolor{codegreen}{rgb}{0.25,0.5,0.25}
\definecolor{codepurple}{rgb}{0.5,0,0.35}
\definecolor{codeblue}{rgb}{0.1,0.1,0.7}
\definecolor{backcolour}{rgb}{0.97,0.97,0.97}
\definecolor{framecolour}{rgb}{0.6,0.6,0.6}
\definecolor{titlebg}{rgb}{0.9,0.9,0.9}
\definecolor{darkred}{RGB}{177, 38, 26}
\definecolor{darkblue}{RGB}{67, 116, 177}
\definecolor{darkgreen}{rgb}{0.0, 0.5, 0.0}
\definecolor{bestcol}{RGB}{  0,102,204} %
\definecolor{goodcol}{RGB}{ 34,139, 34} %

\newcommand{\rowhighlight}{\rowcolor{aliceblue}}

\hypersetup{
  linkcolor = cornellred,
  citecolor  = cadmiumgreen,
  colorlinks = true,
  urlcolor = Blue
}
\usepackage{pifont}
\newcommand{\ours}{\xspace\textbf{LocFac-RL}\xspace }
\newcommand{\longname}{\textbf{Loc}alized and \textbf{Fac}torized  RL with discrete diffusion models}
\newcommand{\dataset}{\textbf{GroundEditReason}}

\definecolor{yuhcol}{rgb}{0.831,0.184,0.494}

\theoremstyle{plain}

\theoremstyle{definition}

\theoremstyle{remark}

\title{Efficient Reinforcement for Visual-Textual Thinking with Discrete Diffusion Model}

\author{%
  Yoonjeon Kim \\
  KAIST\\
  \texttt{yoonkim313@kaist.ac.kr} \\
  \And
  Yuhta Takida \\
  Sony AI \\
  \AND
  Chieh-Hsin Lai \\
  Sony AI \\
  \And
  Eunho Yang \\
  KAIST \& AITRICS 
  \And
  Yuki Mitsufuji \\
  Sony AI \& Sony Group Corporation \\
}

\begin{document}

\maketitle

\begin{abstract}
RL-based post-training has been widely adopted to enable interleaved visual and textual reasoning in unified multimodal models capable of both text and image generation. However, most existing approaches are built upon autoregressive (AR) unified models, which require full image regeneration during visual reasoning. In this work, we demonstrate that multimodal discrete diffusion models are effective alternatives to AR models for reinforcement learning in interleaved reasoning, owing to their ability to perform efficient visual rollouts via \textit{localized visual editing} rather than full image-token regeneration. This reduces rollout computation during GRPO by 26.9\% compared to AR baselines, with minimal performance drop. Despite the improved efficiency, we find that joint reward assignment, which employs a shared reward signal across modalities, introduces cross-modal interference between unrelated image and text token sequences during RL updates. To address this issue, we propose \textit{factorized reward assignment}, a strategy that assigns rewards independently to text and vision segments. With factorized reward assignment, our RL approach achieves an 11.2\% improvement over joint reward assignment and a 38.04\% improvement over the base model.

\end{abstract}

\section{Introduction}
Unified multimodal models have recently gained attention for seamlessly integrating understanding and generation across text and images. Early approaches in unified models largely extended autoregressive (AR) language models into multimodal settings through unified AR architectures \citep{team2024chameleon,tang2025ugen,li2025onecat,chern2024anole,Wu_2025_CVPR,wang2024emu3,yang2025mmar}, or hybrid AR designs \citep{zhou2025transfusion,xie2025showo,Ma_2025_CVPR,pmlr-v235-murty24a,wu2025omnigen2}. More recently, fully discrete diffusion architectures have begun to demonstrate competitive performance in unified multimodal generation and understanding \citep{ai2026llada2,li2025dual,swerdlow2025unidisc,yang2026mmada,wangfudoki,shi2025muddit,li2026lavidao}, suggesting a promising alternative to AR paradigms. While these models unify representation and generation across modalities, an important open question is how they can support complex reasoning processes that unfold across both visual and textual domains.

\begin{figure}[t]
    \centering
    \begin{subfigure}[t]{0.65\textwidth}
        \centering
        \includegraphics[height=0.13\textheight]{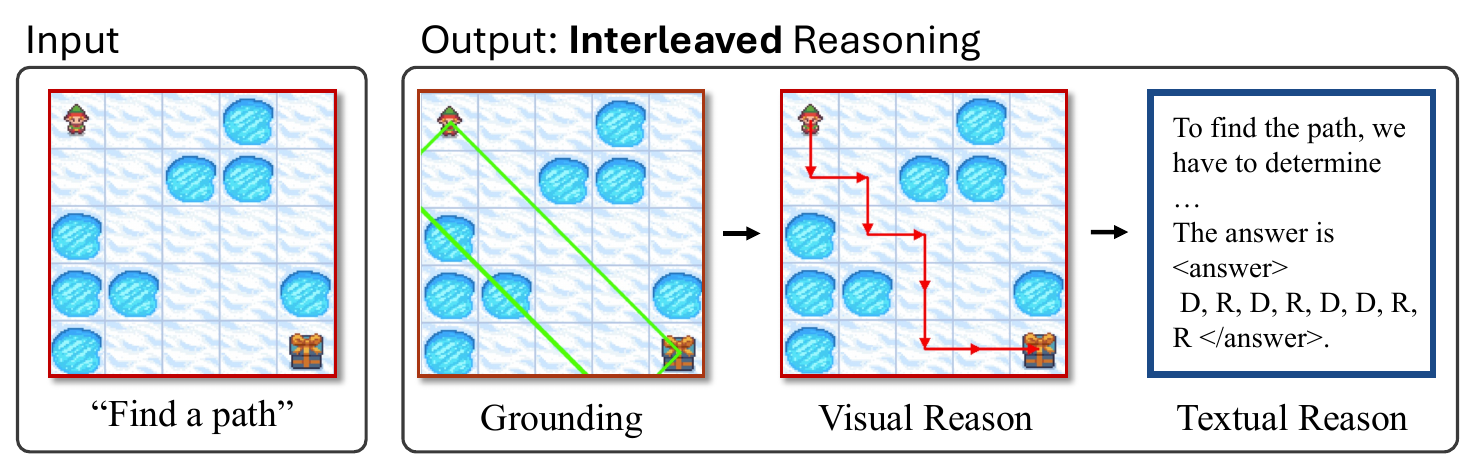}
        \caption{Our goal: textual-visual interleaved reasoning.}
        \label{fig:problem_1}
    \end{subfigure}
    \hfill
    \begin{subfigure}[t]{0.30\textwidth}
        \centering
        \includegraphics[height=0.13\textheight]{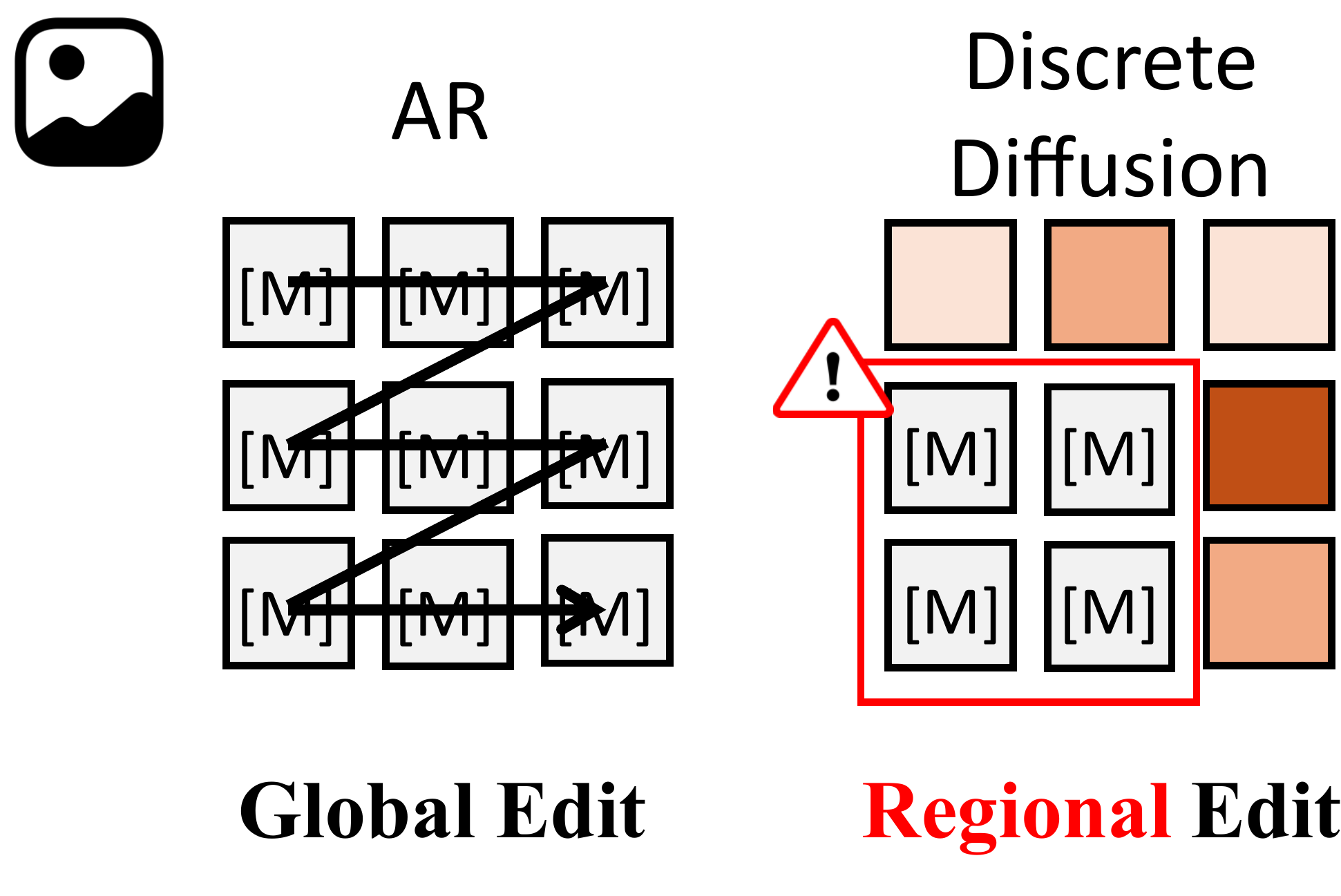}
        \caption{AR vs. Discrete Diffusion.}
        \label{fig:problem_2}
    \end{subfigure}
    
    \caption{(a) Our multimodal model reasons over interleaved textual and visual content within a unified generative framework. (b) AR models should regenerate the entire token sequence even for localized editing due to the nature of the causal decoding. In contrast, discrete diffusion models enable regional editing based on the grounded positions for the visual reasoning, leading to efficient rollouts.}
    \label{fig:problem}
\end{figure}

A key ingredient for such processes is interleaved reasoning, the use of both image states and textual traces throughout the reasoning process, which has proven to outperform text-only reasoning across various multimodal tasks in previous works \citep{gu2026thinkmorph,zhang2026think,cheng2025visual}. For instance, spatial planning requires generating visual signals (e.g. directional arrows), and then continuing reasoning in text conditioned on these updated visual cues as illustrated in \cref{fig:problem_1}. However, despite the strong performance of unified generative models, they lack the ability to perform such interleaved visual–textual reasoning.

To mitigate this limitation, recent works introduce datasets with ground truth intermediate reasoning traces in both image and text \citep{li2025zebra,gu2026thinkmorph}. These datasets provide supervision in the form of chain-of-thought text, alongside predefined transformations on the image. While effective, these approaches confine the model to fixed reasoning patterns dictated by the dataset, such as a limited set of visual operations like zooming, cropping, or overlaying boxes, thereby restricting its capacity to generalize to novel or more flexible reasoning strategies. This motivates the use of reinforcement learning (RL), which enables models to explore diverse reasoning trajectories and optimize them based on task-level rewards rather than fixed supervision \citep{cheng2026omni,xu2026visual}.

Existing approaches that leverage RL on unified generative models remain within AR frameworks \citep{li2025imagine,chern2025thinking,cheng2026omni,xu2026visual}, which necessitates regenerating the entire tokens of the image, often spanning thousands of tokens even when only a small local modification is required, as illustrated in \cref{fig:problem_2}. This leads to a significant computational overhead during RL rollouts, where repeated sampling is essential. 
In this work, we depart from the AR paradigm and propose \textit{localized visual editing}, leveraging the bidirectional context modeling and parallel decoding of discrete diffusion models. This allows the modelsto selectively denoise and update only specific regions from the original image. Concretely, this enables reasoning to proceed through targeted visual modifications with significant acceleration during GRPO \citep{shao2024deepseekmath} rollouts.

However, due to the bidirectional architecture of discrete diffusion models, rewards are assigned jointly over interleaved image and text tokens. Consequently, reward signals intended for one modality (e.g., textual correctness) may inadvertently influence updates in another (e.g., visual tokens), leading to training instability. To address this issue, we propose \textit{modality-factorized reward assignment}, which decouples rewards across modalities and assigns them only to their corresponding token segments.

Our model \ours{}, \longname{}, enhances visual-language interleaved reasoning with following contributions:

\begin{itemize}[itemsep=0mm]
    \item \textbf{Systematization of Multi-modal RL with Discrete Diffusion.} We provide the systematic framework for RL training on multi-modal discrete diffusion models, overcoming the sequential constraints of AR architectures via bidirectional context modeling and parallel decoding.

    \item \textbf{Grounded Localized Visual Rollouts.}
We propose an edit-grounding supervised finetuning stage followed by localized visual editing during the RL phase, significantly reducing the computational cost of visual rollouts while preserving editing effectiveness.
    
    \item \textbf{Modality-Factorized Credit Assignment.} We identify cross-modal credit misassignment under parallel decoding of the discrete diffusion architecture and propose a factorized reward scheme that decouples image and text updates, stabilizing multimodal RL training.
\end{itemize}

\section{Related Works}

\paragraph{RL for Visual-Textual Thinking.}
Recent work leverages the GRPO framework to train vision–language models such as Qwen2.5-VL, which primarily support image understanding through text generation and are augmented with tool-assisted reasoning. Within this framework, prior approaches incorporate external tools such as code generation \citep{zhang2026thyme}, image manipulation modules \citep{zhou2025reinforced,ding2025arm,zheng2026deepeyes}, and textual visualization strategies, which indicates reasoning about visual content through natural language descriptions \citep{huang2026visionr}.

Works that directly leverage image generation into the reasoning process \citep{li2025imagine,chern2025thinking,cheng2026omni,xu2026visual}, often rely on large-scale synthetic SFT data, sometimes combined with GRPO, to encourage coherent multi-step reasoning through generated visual artifacts. These works are primarily built on auto-regressive (AR) or hybrid architectures \citep{chern2024anole,team2024chameleon}, which require token-level sequential decoding. This leads to higher latency and makes localized visual updates inefficient, as modifications require recomputing downstream tokens due to causal dependencies.

\paragraph{Unified Discrete Diffusion Models.}
Unified DDMs generate both images and text through iterative denoising in a discrete token space. Unlike AR models, DDMs allow parallel updates across tokens and naturally support masking-based conditioning. 
Representative models in this multi-modal diffusion models include \citet{li2025dual,swerdlow2025unidisc,yang2026mmada,wangfudoki,shi2025muddit,li2026lavidao,tian2025mmada}. Among these models, LaViDa-O \citep{li2026lavidao} and MMaDA-Parallel \citep{tian2025mmada} are the only open models with image editing ability, which are explicitly trained on large-scale image editing datasets, while others are only capable of text-based image generation.

\paragraph{RL with Multi-Modal Discrete Diffusion Models.}
In multi-modal discrete diffusion models, \citet{ma2026consolidating} introduces a tailored importance estimator and modality-specific rollout strategy to make GRPO feasible despite intractable ratios and \citet{yang2026mmada} introduces a unified policy-gradient framework that jointly masks the output tokens across modalities, using diversified rewards and structured noising strategy.
Extended related works are introduced in \cref{app:extend related works}.

\section{Preliminaries: GRPO with Discrete Diffusion Models}

Let $\pi_\theta$ denote the current policy, and $\pi_{\theta_{\mathrm{old}}}$ the old policy used to generate samples, while $\pi_{\text{ref}}$ is a fixed reference policy. For each prompt $q \sim P(\mathcal{Q})$, the old policy samples a group of $G$ responses, referred to as rollouts, $\mathcal{O} = \{\mathbf{o}_1, \dots, \mathbf{o}_G\}$, each assigned a scalar reward $r_i$.

The GRPO objective maximizes a clipped policy gradient objective with KL regularization, applied at the token level for causal AR models:
\begin{equation}
\resizebox{0.92\linewidth}{!}{$
\begin{aligned}
\mathcal{L}^\text{AR}(\theta)
=
\mathbb{E}_{\substack{q \sim P(\mathcal{Q}) \\ \{\mathbf{o}_i\} \sim \pi_{\theta_{\mathrm{old}}}(\cdot|q)}}
\left[
\frac{1}{G}\sum_{i=1}^G \left(\frac{1}{|\mathbf{o}_i|}
\sum_{t=1}^{|\mathbf{o}_i|}
\left(
\min\!\big(
\rho^\text{AR}_{i,t}(\theta)A_i,\,
\mathrm{clip}_\epsilon(\rho^\text{AR}_{i,t}(\theta))A_i
\big)\right)
- \beta\, D_{\mathrm{KL}}(\pi_\theta(\mathbf o_i\mid q) \,\|\, \pi_{\text{ref}}(\mathbf o_i\mid q) )
\right)
\right],
\end{aligned}
$}
\end{equation}

where the importance sampling ratio at each token $o_{i,t}$ is 
$
\rho^\text{AR}_{i,t}(\theta)
=
\frac{\pi_\theta(o_{i,t}\mid q, \mathbf{o}_{i,<t})}
{\pi_{\theta_{\mathrm{old}}}(o_{i,t}\mid q, \mathbf{o}_{i,<t})},
$
 and $\epsilon$ controls the clipping range between $1-\epsilon$ and $1+\epsilon$. Unlike standard PPO, GRPO assigns a group-normalized advantage shared across all tokens of a response as $A_i
=
\frac{r_i - \mu_{\mathcal{G}}}{\sigma_{\mathcal{G}} + \delta}$,
where $\mu_{\mathcal{G}}$ and $\sigma_{\mathcal{G}}$ are the mean and standard deviation of rewards within the group $\mathcal{G} = \{r_1, \dots, r_G\}$, and $\delta$ is a small constant for numerical stability.

Discrete diffusion models, unlike AR variants, do not have a sequential order. Therefore, following the previous work \citep{tian2025mmada}, the GRPO objective function for discrete diffusion model is defined as
\begin{equation}
\resizebox{0.92\linewidth}{!}{$
\begin{aligned}
\mathcal{L}^\text{DDM}(\theta)
=
\mathbb{E}_{\substack{
q \sim P(\mathcal{Q}) \\
\{\mathbf{o}_i\} \sim \pi_{\theta_{\mathrm{old}}}(\cdot|q)
}}
\left[
    \frac{1}{G}\sum_{i=1}^G 
    \left(
    \frac{1}{|\mathbf{o}_{i,M}|}
    \sum_{k=1}^{|\mathbf{o}_{i,M}|} 
    \left(
    \min\!\big(
    \rho^\text{DDM}_{i, k}(\theta)A_i,\,
    \mathrm{clip}_\epsilon(\rho^\text{DDM}_{i,k}(\theta))A_i
    \big)
    \right)
    - \beta\, D_{\mathrm{KL}}\left(\pi_\theta(\mathbf{o}_i\mid q) \,\|\, \pi_{\text{ref}}(\mathbf{o}_i\mid q)\right)
    \right)
\right].
\end{aligned}
$}
\end{equation}
 
Here, the importance sampling ratio for discrete diffusion models, $\rho^\text{DDM}_{i,k}(\theta)$, is defined over a masked token $o_{i,k} \in \mathbf{o}_{i,M}$, conditioned on the prompt $q$ and the unmasked sequence $\mathbf o_{i,\overline M}$. More specifically, $M$ is a binary mask over the sequence $\mathbf o_i$, while $\overline M$ denotes the complement. 

Formally, importance sampling ratio for the masked token $o_{i, k} \in \mathbf{o}_{i,M}$ is defined as 
$
\rho_{i,k}^\text{DDM}(\theta)
=
\frac{\pi_\theta(o_{i,k} \mid q,  \mathbf o_{i,\overline M})}{\pi_{\theta_{\mathrm{old}}}(o_{i,k} \mid q, \mathbf o_{i,\overline M})}.
$

\begin{figure}[t]
    \centering

    \begin{subfigure}[t]{0.28\textwidth}
        \centering
        \includegraphics[height=0.23\textheight]{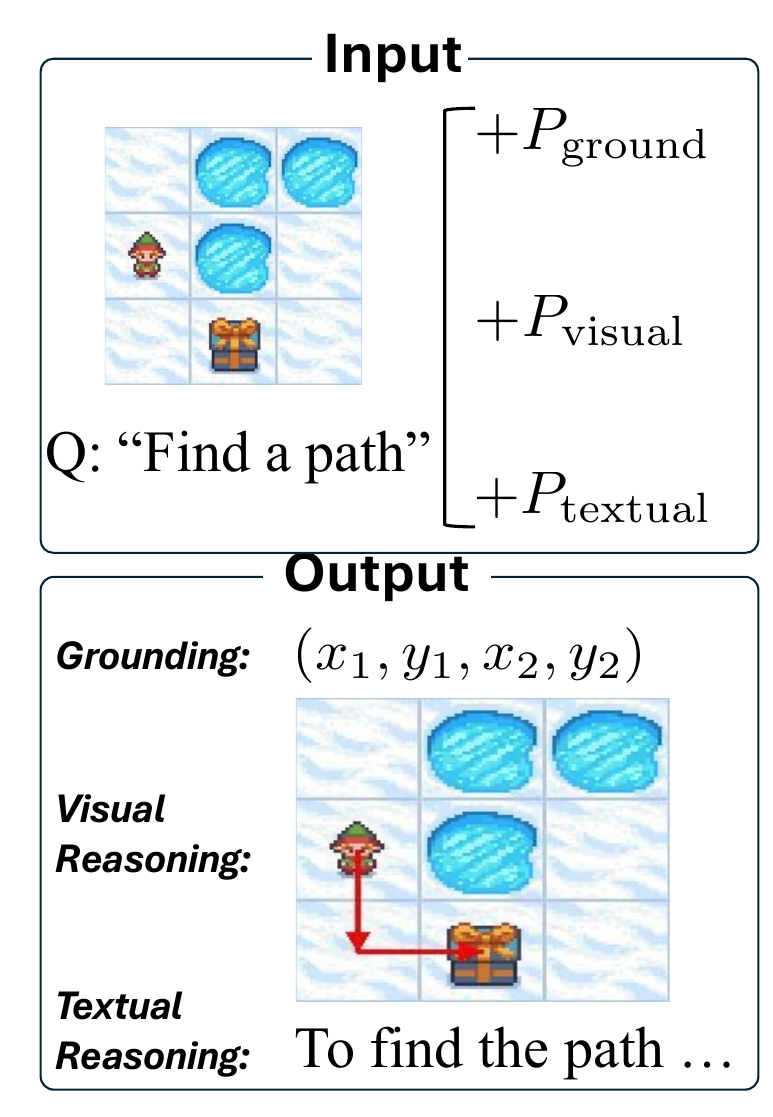}
        \caption{\textbf{SFT phase.}}
    \end{subfigure}
    \hfill
    \begin{subfigure}[t]{0.70\textwidth}
        \centering
        \includegraphics[height=0.22\textheight]{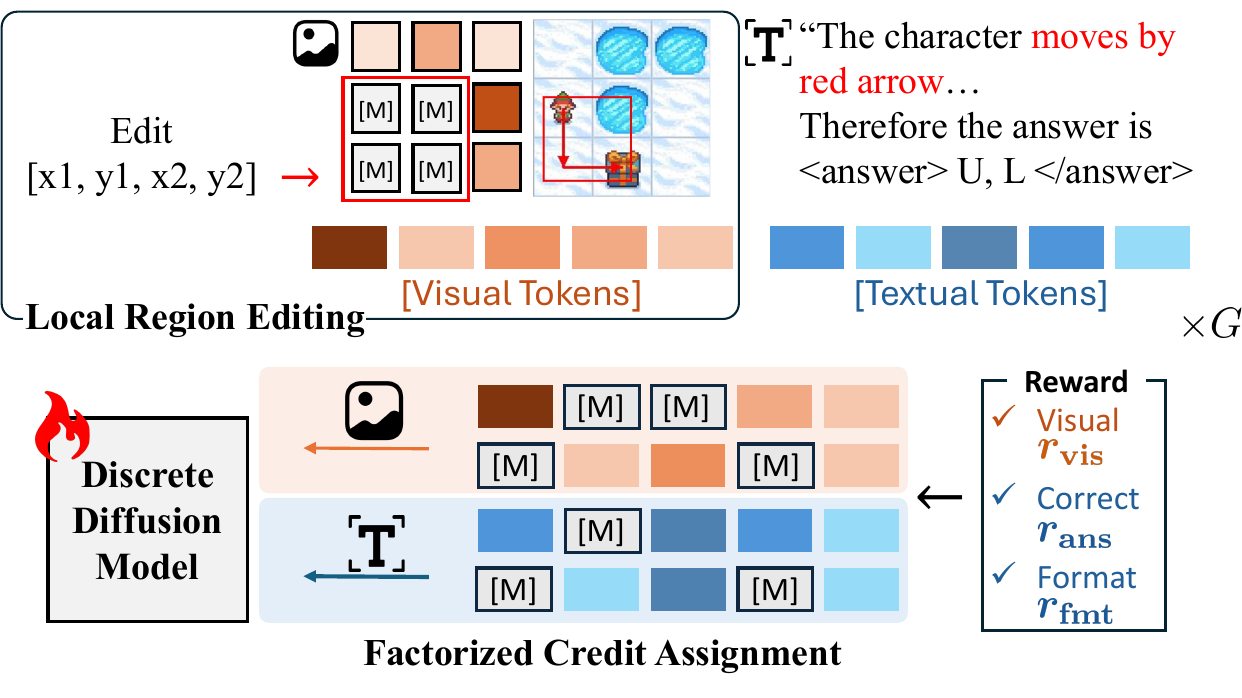}
        \caption{\textbf{GRPO phase.}}
    \end{subfigure}

    \caption{Overview of the proposed training framework consisting of SFT and GRPO phases. (a) During SFT, the model is trained to generate grounding, visual reasoning, and textual reasoning in an interleaved input–output format. (b) During GRPO, the model performs multimodal reasoning by sequentially generating: i) grounding regions that require editing, ii) localized visual edits conditioned on the image context, and iii) textual reasoning based on the edited visual content. The resulting multimodal reasoning sequence is used to update the policy model in a \textit{factorized} manner.}
    \label{fig:proposed}
\end{figure}

\section{Methodology}
We introduce \ours{}, a two-phase framework consisting of SFT and \longname{}. Within this framework, the model operates on an input query comprising a problem image and its accompanying text. It performs multimodal reasoning by first generating a visual cue from the image, followed by subsequent textual reasoning.

To effectively learn this behavior, we adopt a two stage training procedure (\cref{subsec:training}). We perform supervised fine-tuning (SFT) to initialize the model with multimodal chain-of-thought capabilities and grounding for localized image editing. Then, we proceed to reinforcement learning (RL) to optimize the model across diverse multi-modal reasoning tasks. Our RL strategy is tailored for multimodal discrete diffusion and introduces two core mechanisms (\cref{subsec:local_edit}): i) accelerated visual rollouts achieved through localized, region-specific denoising, and ii) factorized modality-specific credit assignment under the GRPO objective.

\paragraph{Problem Formulation.}
The input comprises an image and a text query $(I_q, T_q)$. The model $\pi_\theta$ generates an intermediate reasoning trace $I_\text{gen}, T_\text{gen}$ in token space, where visual sequence ($I_\text{gen}$) is synthesized prior to the textual sequence ($T_\text{gen}$). For localized image editing, the model predicts a bounding box in [LOC] format to identify specific target regions within $I_q$.

To enable this reasoning process, the model is prompted to perform three sequential actions using templates for grounding ($P_\text{ground}$), visual reasoning ($P_\text{visual}$), and textual reasoning ($P_\text{textual}$). The full prompt template is in the appendix \cref{app: prompt}.

\subsection{Overall Framework: Cold-Start SFT and RL}\label{subsec:training}

\paragraph{SFT Phase.} To equip the base model with long chain-of-thought reasoning in a multimodal, interleaved setting and to enable accurately grounded, localized editing during the RL phase, we train it on \dataset{}, an augmented dataset which is constructed from \textit{ZebraCoT} \citep{li2025zebra} and  \textit{ThinkMorph} \citep{gu2026thinkmorph}. We automatically annotate regions corresponding to intermediate modifications by comparing each reasoning image with its source image, since the original dataset lacks explicit spatial supervision for intermediate editing steps. As shown in \cref{fig:problem_1}, the dataset consists of input questions $(I_q, T_q)$, its ground truth reasoning traces $(I_\text{gt}, T_\text{gt})$, and the bounding box $B_\text{gt}$ that indicates the region of difference between question image $I_q$ and the reasoning image $I_\text{gt}$. The augmentation process and the dataset sample is shown in the appendix \cref{app: dataset construction}. 

To jointly train the model for spatial grounding and multimodal reasoning, we adopt a unified masked reconstruction objective under a discrete diffusion framework. Specifically, we construct task-dependent prompt–response pairs $(p_0, r_0)$ from the augmented dataset as 
\[(p_0, r_0) =\begin{cases}([I_q, T_q, P_\text{ground}],\; B_\text{gt}) & \text{for grounding}, \\([I_q, T_q, P_\text{visual}, P_\text{textual}],\; [I_\text{gt}, T_\text{gt}]) & \text{for reasoning}.\end{cases}\]

Given $r_0$, we sample a timestep $t$ and construct a corrupted sequence $r_t$ by randomly masking tokens. The model is then trained to reconstruct the masked tokens conditioned on both the prompt $p_0$ and the partially observed sequence $r_t$. The unified supervised fine-tuning objective is defined as:
\[\mathcal{L}_{\text{SFT}} =- \mathbb{E}_{t,\, p_0,\, r_0,\, r_t}\left[\frac{1}{t} \sum_{i=1}^{|r_0|}\mathbb{I}\big[r_t^i = [\text{MASK}]\big]\;\log p_\theta\big(r_0^i \mid p_0, r_t\big)\right]\]
where $x_0=[p_0;r_0]$, $x_t=[p_0;r_t]$ so SFT is equivalent to masked denoising pretraining with masking applied only to $r_0$ and the ratio of ($p_0$, $r_0$) sampling between grounding and reasoning is set to 0.5.

\paragraph{GRPO Phase.} Following the SFT phase, the model develops grounded multimodal reasoning capabilities, serving as a strong initialization for the subsequent reinforcement learning (RL) stage. However, SFT is constrained by the limited scale and diversity of available multimodal reasoning annotations, which are costly to obtain. To address this, we further optimize the model via RL on a broader and more diverse set of tasks. Specifically, we use a held-out split of \textit{ThinkMorph} comprising 4K examples not seen during SFT, encouraging the model to generalize beyond the original training data. We additionally incorporate 1K examples from \textit{ArxivQA} \citep{li2024multimodal}, yielding a total of 5K training instances spanning visual reasoning, perception, math and science domains following previous works \citep{cheng2026omni,zhao2026thinking}.

We optimize the policy $\pi_\theta$ using GRPO with rewards defined over visual output $I_\text{gen}$ and textual output $T_\text{gen}$. The visual reward is defined as a CLIP similarity,
$
r_{\text{vis}} = \texttt{CLIP}(I_\text{gen}, I_\text{gt}) \in [0,1],
$
between the generated image $I_\text{gen}$ and the ground truth reasoning image $I_\text{gt}$. This reward is continuous-valued, providing a smooth signal for optimization.

The textual reward consists of answer and format reward. The answer reward is formulated as
$
r_{\text{ans}} =
\mathbb{I}\!\left[\mathrm{extract}({T}_{\text{gen}})=a^\star\right] \in \{0, 1\},
$
where $a^\star$ is the answer and $\mathrm{extract}(\cdot)$ parses the prediction that is structured with required output format from the generation $T_\text{gen}$. Format reward $r_{\text{fmt}} \in \{0,1\}$ enforces correct answer formatting within \texttt{<answer> </answer>} tags. The rewards are aggregated by $r_i = r_{i,\text{ans}} + r_{i,\text{vis}} + r_{i,\text{fmt}}/2$ and then normalized per group.

Also, note that \textit{ArxivQA} do not include ground truth reasoning images, thus the visual reward is dropped for such cases, allowing the policy to explore visual edits based solely on downstream task rewards rather than explicit image-level supervision.

\subsection{\ours{}: Localized Edit for Visual Rollout and Factorized Credit Assignment}\label{subsec:local_edit} 

\paragraph{Localized Editing in Visual Rollout for Accelerated GRPO.} A key challenge in the RL stage is the computational cost of visual rollouts during GRPO training, where full-image generation incurs substantial denoising overhead. To address this, we leverage the architectural advantages of bidirectional discrete diffusion models to enable localized image editing conditioned on the predicted grounding regions. Instead of regenerating the entire image, the model selectively refines only the regions specified by the grounding signal, while preserving the remaining context. This localized editing mechanism—where only a small subset of image tokens is masked—substantially reduces the number of denoising steps required, thereby improving training efficiency without compromising fidelity. This fundamentally differs from AR models which necessitate global editing due to their causal structure, where changes to earlier tokens propagate and affect all subsequent tokens. 

Given an input image and text, the model predicts a grounding box defined by the coordinates $(x_1, y_1, x_2, y_2)$ in pixel space of $(H \times W)$, which is downsampled onto the VQ-latent grid of shape $H_\text{VQ} \times W_\text{VQ}$. Then, the VQ tokens within the specified region are replaced with the \texttt{[MASK]} token for editing. In contrast, tokens outside this region are preserved by directly copying them from the original image. This approach restricts the denoising objective to a localized subset of the latent space, significantly reducing the computational overhead of the generation process. Consequently, the effective computational cost of the diffusion reverse steps is scaled from the original $K$ steps to a fractional equivalent:
$\frac{K \cdot (x_2 - x_1)(y_2 - y_1) \cdot(H_{\text{VQ}} \times W_{\text{VQ}})}{H \times W}$. This localized denoising strategy ensures that visual rollouts are both computationally efficient and spatially consistent with the original context.

\paragraph{Factorized Policy for Stable Credit Assignment.} 

While localized editing resolves the computational bottlenecks of RL rollouts, using discrete diffusion models introduces another challenge. We show that, unlike AR models, which have an inherent sequential ordering, masked discrete diffusion models require modality factorization to prevent content leakage from future text segments into earlier image segments. 

More formally,
let $X=[X_I,X_T]$ denote a sequentially generated multimodal output where $X_I$ is visual and $X_T$ is textual reasoning. Then, the reward $R(X)$ over $X$ is decomposed as $R(X)=R_1(X_I)+R_2(X_I,X_T)$, 
where $R_1(X_I)=r_\text{vis}$ and 
$R_2(X_I,X_T)=r_\text{fmt}+r_\text{ans}/2$\footnote{Note that $R_1$ is measured solely based on the image generation, while $R_2$ is rewarded by the final answer format and correctness which conditions on the earlier image generation.}. 

Under KL-regularized policy optimization, the optimal update for policy model $\pi_k$ at 
optimization step $k$ given prompt $q$ is given by
$
\pi_{k+1}(X\mid q)\propto \pi_k(X\mid q)\exp\!\left(\beta^{-1}R(X)\right),
$
as derived in \citep{liu2019neural}.
In AR models, the policy can be marginalized for $X_I$, sequentially 
following the generation order as 
$\pi_k(X\mid q)=\pi_k(X_I\mid q)\,\pi_k(X_T\mid q,X_I)$. Factorizing and marginalizing over the future segment $X_T$ yields the 
exact marginal update for $X_I$:
\begin{equation}\label{eq:ar_image_seq}
\pi_{k+1}(X_I\mid q)\propto \pi_k(X_I\mid q)\exp\!\left(\beta^{-1}R_1(X_I)\right)\,\mathbb{E}_{\pi_k(X_T\mid q,X_I)}\!\left[\exp\!\left(\beta^{-1}R_2(X_I,X_T)\right)\right].
\end{equation}

Analogously, masked discrete diffusion models (DDMs) define denoising
conditionals over masked tokens, $X_M=[X_{I,M},X_{T,M}]$, given their unmasked
complements, $X_{\overline{M}}=[X_{I,\overline{M}},X_{T,\overline{M}}]$.
Conditioning on $X_{\overline{M}}$ and marginalizing the
remaining masked text tokens yields the exact conditional update for the
image segment:
\begin{equation}
\label{eq:ddm_image_cond}
\resizebox{\linewidth}{!}{$
\pi_{k+1}(X_{I,M}\mid q, X_{\overline{M}})
\propto
\pi_k(X_{I,M}\mid q, X_{\overline{M}})\,
\exp\!\left(\beta^{-1}R_1(X_I)\right)
\mathbb{E}_{\pi_k(X_{T,M}\mid q,\,X_{T,\overline{M}},\,X_I)}
\!\left[
\exp\!\left(\beta^{-1}R_2(X_I,X_T)\right)
\right],
$}
\end{equation}
which integrates over the future text segment under image-first generation where $X_{T,\overline{M}}=\emptyset$.
Under such regime, masked views constructed from over whole sequence, $[X_I, X_T]$, may unmask the future text tokens while image tokens remain
masked, resulting in causal leakage.

To eliminate this mismatch, we restrict the conditioning context of the
image update to the unmasked image tokens, removing the dependence
on masked text tokens $X_{T,\overline{M}}$, and derive modality-wise updates by freezing
the complementary modality. For the image segment, we update the policy
under the inference-consistent context using the visual reward:
\begin{equation}\label{eq:ddm_image_update}
\pi_{k+1}(X_{I,M}\mid q, X_{I,\overline{M}})
\propto
\pi_k(X_{I,M}\mid q, X_{I,\overline{M}})
\exp\!\left(\beta^{-1}R_1(X_I)\right).
\end{equation}
For the joint reward $R_2(X_I,X_T)$, we condition on the completed image
segment $X_I$ and update the text conditionals:
\begin{equation}\label{eq:ddm_text_update}
\pi_{k+1}(X_{T,M}\mid q, X_I, X_{T,\overline{M}})
\propto
\pi_k(X_{T,M}\mid q, X_I, X_{T,\overline{M}})
\exp\!\left(\beta^{-1}R_2(X_I,X_T)\right).
\end{equation}

\section{Experiments}\label{sec:experiments}
In this section, we provide the details of training and evaluation configuration in \cref{subsec:eval detail}. We demonstrate the performance gain driven by SFT and RL of \ours{} and the efficiency gain of our method through experiments in \cref{subsec: analysis}. Finally, we demonstrate component-wise studies on vision-language modalities, grounding ability, and visual reasoning ability in \cref{subsec:component analysis}.

\subsection{Experimental setup}\label{subsec:eval detail}

\paragraph{Models.}  We evaluate our method on two architectures: MMaDA-Parallel and LaViDa-O. MMaDA-Parallel employs a unified multimodal diffusion language model, where text and image tokens are co-denoised in parallel within a shared backbone. In contrast, LaViDa-O uses a masked diffusion framework with an Elastic Mixture-of-Transformers, separating understanding and generation branches. Among open-source multimodal discrete diffusion models, only LaViDa-O and MMaDA-Parallel support image editing, while MMaDA and Mmudit are limited to text-to-image generation without editing capabilities. Following prior work \citep{cheng2026omni}, we adopt ANOLE as an AR baseline.

\paragraph{Evaluation Dataset.}
We evaluate our method on MM-Vet, MMMU (validation), V*Bench, CVBench, ChartQA, and BLINK-Jigsaw using the LMMs-Eval package, consistent with the evaluation protocol of our baseline model, LaViDa-O. 

\paragraph{Evaluation Setting.}
Across all models, generation follows an interleaved reasoning format: image editing is performed first, followed by textual reasoning conditioned on the generated visual and textual context. Evaluation parameters are set to a temperature of 0.0 following the hyperparameter settings of the baseline model, with maximum generation lengths tailored to each task as detailed in the appendix \cref{tab:eval_token_length}. For benchmarks requiring over 1024 tokens—specifically MMVet, CVBench, and BLINK—we utilize Chain-of-Thought (CoT) prompting with task-specific default templates. Performance is measured via exact-match accuracy. For MM-Vet, which provides ground-truth annotations in an A <OR> B format, a prediction is deemed correct if the extracted response matches either valid option.
Note that all training and evaluation hyper-parameters are specified in appendix \cref{app: Training Details,app: evaluation details}.

\subsection{Analysis}\label{subsec: analysis}

\paragraph{Effect of SFT and RL phase.}
As shown in \cref{tab:main}, baseline performance improves substantially across the SFT and RL stages, with SFT providing consistent initial gains and subsequent RL further enhancing accuracy. We omit results for RL without SFT phase because the model fails to reliably follow the structured output formats required for reward extraction and lacks a stable starting policy for the complex reasoning and image generation tasks necessary for effective reinforcement learning.
The relatively low initial performance of the MMaDA-Parallel model is expected, as it is fine-tuned from the base MMaDA model exclusively on image generation and editing data, without exposure to textual reasoning traces for image understanding. After SFT, basic image understanding capabilities are effectively recovered, providing a robust initialization point for the subsequent RL training.

\begin{table}[t]
\small
\setlength{\extrarowheight}{3pt}
\centering
\caption{Performance comparison on multimodal benchmarks. RL in this table refers to GRPO with \textit{global} editing for fair comparison with the AR baseline. Scores are reported in accuracy (\%). Green percentage indicates the reduction in per-step train time of \ours{} compared to the global version.}
\label{tab:main}
\resizebox{\linewidth}{!}{
\begin{tabular}{l l c c c c c c | c}
\hline
\rowhighlight
\textbf{Arch} & \textbf{Model} & MMVeT & MMMU (val) & VStarBench & ChartQA & BLINK Jigsaw & CVBench & \textbf{Average Accuracy} \\
\hline

\multirow[c]{3}{*}{\textbf{\raisebox{-1.2ex}{AR}}}
& \textbf{ANOLE} & 7.8 & 2.6 & 7.3 & \textbf{7.3} & 28.0 & 14.4 & 11.23 \\
\cmidrule(l{3pt}r{3pt}){2-9}
& \quad + SFT & 7.8 & 20.2 & 18.3 & 5.8 & 36.7 & 17.6 & 17.73 \\
& \cellcolor{gray!15}\quad + RL & \cellcolor{gray!15}\textbf{9.6} & \cellcolor{gray!15}\textbf{21.0} & \cellcolor{gray!15}\textbf{20.9} & \cellcolor{gray!15}5.7 & \cellcolor{gray!15}\textbf{38.7} & \cellcolor{gray!15}\textbf{18.4} & \cellcolor{gray!15}\textbf{19.05} \\
\midrule

\multirow[c]{6}{*}{\textbf{\raisebox{-1.2ex}{D-Diff}}}
& \textbf{LaViDa-O} & 9.36 & \textbf{43.33} & 19.90 & 38.68 & 44.77 & {59.70} & 35.96 \\
\cmidrule(l{3pt}r{3pt}){2-9}

& \quad + SFT & 11.93 & {43.11} & {31.94} & 19.64 & {43.61} & 48.67 & 33.15 \\

& \cellcolor{gray!15}\quad + RL (Global + Factorized)
& \cellcolor{gray!15}{19.71} 
& \cellcolor{gray!15}\textbf{43.78} 
& \cellcolor{gray!15}\textbf{37.17} 
& \cellcolor{gray!15}\textbf{66.16} 
& \cellcolor{gray!15}43.48 
& \cellcolor{gray!15}\textbf{64.25} 
& \cellcolor{gray!15}\textbf{45.76} \\

& \cellcolor{gray!15}\quad + RL (\ours{}) & 
\cellcolor{gray!15}\textbf{20.64} & 
\cellcolor{gray!15}42.78 & 
\cellcolor{gray!15}29.32 & 
\cellcolor{gray!15}{63.36} & 
\cellcolor{gray!15}\textbf{45.43} & 
\cellcolor{gray!15}{51.90} & 
\cellcolor{gray!15}{42.24} \textcolor{codegreen}{\textbf{($\downarrow$ 12.3\% s/step)}} \\
\cmidrule(l{3pt}r{3pt}){2-9}

& \textbf{MMaDA-Parallel} & 5.05 & 0.00 & 0.00 & 0.00 & 0.68 & 3.87 & 1.60 \\
\cmidrule(l{3pt}r{3pt}){2-9}

& \quad + SFT & 9.17 & {16.11} & 26.70 & 3.64 & 47.33 & {36.81} & 23.29 \\
& \cellcolor{gray!15}\quad + RL (Global + Factorized)
& \cellcolor{gray!15}\textbf{12.84} 
& \cellcolor{gray!15}\textbf{17.22} 
& \cellcolor{gray!15}\textbf{30.37} 
& \cellcolor{gray!15}{3.96} 
& \cellcolor{gray!15}{52.00} 
& \cellcolor{gray!15}36.54 
& \cellcolor{gray!15}{25.49} \\
& \cellcolor{gray!15}\quad + RL (\ours{}) & 
\cellcolor{gray!15}\textbf{12.84} & 
\cellcolor{gray!15}15.44 & 
\cellcolor{gray!15}{28.27} & 
\cellcolor{gray!15}\textbf{6.52} & 
\cellcolor{gray!15}\textbf{55.33} & 
\cellcolor{gray!15}\textbf{37.98} & 
\cellcolor{gray!15}\textbf{26.06} \textcolor{codegreen}{\textbf{($\downarrow$ 16.9\% s/step)}} \\
\bottomrule
\end{tabular}
}
\end{table}

\paragraph{Local Editing vs. Global Editing within Discrete Diffusion Model}

Localized editing provides consistent efficiency improvements over global editing, reducing the average visual rollout time by $26.9\%$ for LaViDa-O and $52.5\%$ for MMaDA-Parallel in \cref{tab:combined}. This also translates to overall training time savings of $12.3\%$ and $16.9\%$, respectively, highlighting its practical advantage in reducing computational cost. Importantly, as shown in \Cref{tab:main}, these efficiency gains using local editing are achieved while maintaining the performance compared to the RL with global editing and factorized credit assignment.

We observe modest accuracy reductions on VStarBench and CVBench, suggesting that a some tasks benefit from the broader updates enabled by global editing. Despite the grounding performance achieving high IoU score as shown in \cref{tab:grnd visual}, there are cases where grounding error leads to compromise in local editing. We provide the failure cases and analysis in appendix \cref{app:qualitative}. Nevertheless, the overall results indicate that localized editing strikes a favorable balance between efficiency and performance. For MMaDA-Parallel, which is not trained on grounding corpus, therefore do not have an ability to predict bounding box regions, we use the ground truth bounding box for localized editing. As a result, localized editing slightly outperforms global editing in several benchmarks as it allows targeting editing on necessary regions, further underscoring its effectiveness.

\paragraph{AR Model vs. Discrete Diffusion Model}
The training time reductions for the AR model, ANOLE-7B, and the discrete diffusion model, LaViDa-O 8B, are not directly comparable due to inherent differences in model architecture, parameter scale, and inference strategies. Nevertheless, our evaluation highlights the efficiency of discrete diffusion models which enables localized image editing. The total train time reduces by $79.6\%$ comparing the discrete diffusion model LaViDa-O (1041 sec/step) and the AR model ANOLE (2489 sec/step), which inevitably requires regeneration of the entire image token sequence. This global approach becomes prohibitively inefficient during the reinforcement learning phase due to the resulting elongation of the image rollout process. These findings demonstrate that the localized editing capability of diffusion-based models provides a scalable advantage for multi-modal reinforcement learning over traditional auto-regressive methods.

\begin{table}[t]
\small
\centering
\caption{Efficiency and performance comparison across AR and diffusion-based architectures, including global vs.\ localized editing settings. Scores are reported in accuracy (\%). $\Delta $RL indicates the accuracy gain from the SFT model. }
\label{tab:combined}
\centering
\caption{Efficiency comparison}
\resizebox{\linewidth}{!}{
\begin{tabular}{llcccc}
\rowhighlight
\hline
\textbf{Model} & \textbf{Editing Mode} & \textbf{Visual Rollout (s/step)} & \textbf{Train Time (s/step)} & \textbf{Avg. Acc.} & \textbf{$\Delta$ RL} \\
\hline
ANOLE-7B & Global (AR) & 1982 & 2487 & 19.05 & +7.44\% \\
\midrule
\multirow{2}{*}{LaViDa-O 8B} & Global & 552 & 1187 & \textbf{45.76} & +38.04\%\\
 & \textbf{Localized} & \textbf{404 ($\downarrow 26.9\%$)} & \textbf{1041 ($\downarrow 12.3\%$)} & 42.24 & +27.42\% \\
\midrule
\multirow{2}{*}{MMaDA-Parallel 8B} & Global & 362 & 1221 & 25.49 & +9.45\%\\
 & \textbf{Localized} & \textbf{172 ($\downarrow 52.5\%$)} & \textbf{1012 ($\downarrow 16.9\%$)} & \textbf{26.06}  & +11.89\% \\
\bottomrule
\end{tabular}
}
\end{table}

\subsection{Ablation Studies}\label{subsec:component analysis}
\paragraph{Uni-Modal RL vs. Multi-Modal RL}
We evaluate the impact of multi-modal integration during the reinforcement learning phase by comparing our approach against uni-modal baselines in \cref{tab:modality_ablation}. These baselines generate either text or images in isolation and optimize the model based on modality-specific rewards such as text-based accuracy or CLIP-based visual similarity. In contrast, our method unifies both modalities through interleaved reasoning during the rollout and update phases. Experimental results demonstrate that this unified approach yields superior performance across most of the evaluation benchmarks. This suggests that multi-modal optimization facilitates a more cohesive alignment between visual evidence and linguistic reasoning than can be achieved through single-modality training.

\begin{table}[bh]
\small
\centering
\caption{Ablation on Modality during Reinforcement Learning Training. Scores reported in accuracy (\%).}
\label{tab:modality_ablation}
\resizebox{\linewidth}{!}{
\begin{tabular}{llcccccc|c}
\hline
\rowhighlight
\textbf{Model} & \textbf{RL Variant} & MMVeT & MMMU (val) & VStarBench & ChartQA & BLINK Jigsaw & CVBench & \textbf{Average Accuracy} \\
\hline
\multirow{3}{*}{LaViDa-O} 
    & Image Only RL & 18.35 &  {42.78} & 3.60  & 60.20 &  \textbf{45.75} & {53.18} & 37.31 \\
    & Text Only RL  & 17.43 & 42.60  & 36.13 & 57.24 &  \textbf{45.75} & {53.34} & 42.08 \\
    & Both (\ours{})    & \textbf{19.71} & \textbf{43.78}  & \textbf{37.17} & \textbf{66.16} & {43.48} & \textbf{64.25} & \textbf{45.76} \\
\midrule
\multirow{3}{*}{MMaDA-Parallel} 
    & Image Only RL & 9.63 & 14.78 &  26.18 & 3.60 & 48.00 & \textbf{38.17} & 23.39 \\
    & Text Only RL  & 11.93 &  15.33 &  26.18 & 3.84 & \textbf{55.33} & 36.24 & 24.81 \\
    & Both (\ours{})   & \textbf{12.84} & \textbf{17.22} & \textbf{30.37} & \textbf{3.96} &  {52.00} & 36.54 & \textbf{25.49} \\
\bottomrule
\end{tabular}
}
\end{table}

\begin{table}[th]
\small
\centering
\caption{Ablation on Credit Assignment across Modalities. The factorized credit assignment shows overall stronger performance across six benchmarkts, compared to joint credit assignment. Scores reported in accuracy (\%).}
\label{tab:separate_unify}
\resizebox{\linewidth}{!}{
\begin{tabular}{llcccccc|c}
\rowhighlight
\hline
&  & MMVeT & MMMU (val) & VStarBench & ChartQA & BLINK Jigsaw & CVBench & \textbf{Average Accuracy} \\
\hline
\multirow{2}{*}{LaViDa-O} & Joint  & {18.81} & {42.56} & 33.84 & {62.40} & \textbf{45.25} & 44.05 & 41.15 \\
& \textbf{Factorized} & \textbf{19.71} & \textbf{43.78} & \textbf{37.17} & \textbf{66.16} & {43.48} & \textbf{64.25} & \textbf{45.76} \textcolor{codegreen}{\textbf{(+11.2\%)}}\\
\midrule
\multirow{2}{*}{MMaDA-Parallel} & Joint & 10.09 & \textbf{17.22} & \textbf{30.37} & \textbf{4.44} &  47.67 & 35.14 & 24.16\\
& \textbf{Factorized} & \textbf{12.84} & \textbf{17.22} & \textbf{30.37} & {3.96} &  \textbf{52.00} & \textbf{36.54} & \textbf{25.49} \textcolor{codegreen}{\textbf{(+5.5\%)}} \\
\bottomrule
\end{tabular}
}
\end{table}

\paragraph{Ablation of Factorized Credit Assignment.}
We observe in \cref{tab:separate_unify} that using joint credit assignment over the text and image tokens leads to overall performance drop for both LaViDa-O and MMaDA-Parallel. This proves that there exists a spurious credit assignment within a joint credit assignment.

\paragraph{Grounding Accuracy Analysis}
Following the Supervised Fine-Tuning (SFT) phase, we observe a marked improvement in the alignment between predicted and ground-truth bounding boxes when identifying localized regions for editing. This spatial precision is critical for the success of region-specific denoising. To quantify this progress, we evaluate the Intersection over Union (IoU) scores for LaViDa-O, MMaDA-Parallel, and Anole-7B. As shown in \cref{tab:grnd visual}, the substantial increase in IoU across all architectures demonstrates the effectiveness of SFT on our augmented dataset. These results suggest that the fine-tuning process successfully calibrates the model to translate textual reasoning into accurate spatial coordinates, providing a robust foundation for the subsequent localized rollout phase.

\begin{wraptable}[10]{r}{0.4\linewidth}
\vspace{-11pt}
\centering
\small
\caption{Grounding accuracy (IoU Score), and visual reasoning fidelity (CLIP similarity) using ThinkMorph Validation Set.}
\label{tab:grnd visual}
\resizebox{\linewidth}{!}{%
\begin{tabular}{lcc}
\toprule
& \makecell[c]{\textbf{Grounding} \\ \textbf{(IoU)}} & \makecell[c]{\textbf{Visual Reasoning} \\ \textbf{(CLIP sim)}} \\
\midrule
Base model & 0.384 & 0.637 \\
\midrule
+ SFT & 0.716 & 0.825 \\
+ \ours{} & \textbf{0.826} & \textbf{0.831} \\
\bottomrule
\end{tabular}%
}
\end{wraptable}

\paragraph{Visual Reasoning and Fidelity Analysis}
To evaluate the enhancement of visual reasoning capabilities following the RL stage, we assess the CLIP similarity (ViT-L/14) between the model-edited images and the ground-truth reasoning frames. This analysis is conducted using the ZebraCoT and ThinkMorph validation sets, both of which are strictly held out from the SFT and RL training phases to ensure an unbiased measure of generalization. As reported in \cref{tab:grnd visual}, the metrics indicate that the visual reasoning ability improves after both SFT and RL stage.

\section{Limitations}
Our framework presents several limitations that offer several directions for future research. Currently, it operates within a discrete modality space, excluding the high-fidelity synthesis of continuous latent or diffusion-based models. Future work could bridge this gap by using the discrete engine as a controller for diffusion decoders. Additionally, the RL stage is constrained by SFT initialization and reward signal sensitivity, which could be addressed through training with robust cold-start SFT dataset, and implementing a robust reward on both image and text modalities to stabilize the exploration. Finally, while our localized editing strategy ensures precision, it is not currently optimized for global transformations like zooming or novel viewpoint synthesis. While this regional focus does not inherently degrade performance on global tasks, it lacks the holistic pixel-shifting required for them.

\section{Conclusion}
We have introduced \ours{}, a unified framework for interleaved visual-language reasoning that enables efficient learning across text and visual domains. Our approach aims to equip models with the ability to generate, refine, and ground intertwined textual and visual outputs in open-ended environments. Through supervised fine-tuning for interleaved reasoning, a reinforcement learning framework for general applicability, accelerated rollout via localized editing, and a masked discrete diffusion language modeling strategy for unified credit assignment, our method significantly improves both the scalability and effectiveness of multimodal learning, addressing key limitations of auto-regressive baselines and costly rollout generation.

\clearpage
\bibliography{bib/list}
\bibliographystyle{bib/format}
\clearpage
\appendix
\section{Extended Related Works}\label{app:extend related works}

\subsection{Unified Multimodal Generative Models}

\paragraph{Auto-regressive models.}
AR models treat multi-modal data as a single sequence and generate tokens causally. Early approaches rely on discrete visual tokens via vector quantization (e.g., Chameleon \citep{team2024chameleon}, Anole \citep{chern2024anole}, Janus \citep{Wu_2025_CVPR}, Emu \citep{wang2024emu3}). More recent works explore different visual representations: discrete-token models such as UGen \citep{tang2025ugen} and OneCAT \citep{li2025onecat} unify text and image tokens in a shared vocabulary, while continuous-token models such as UniFluid \citep{fan2025unified} and OmniGen-AR \citep{wu2025omnigen2} remove the quantization bottleneck by directly modeling continuous latent embeddings.

A key limitation of AR models lies in their causal structure. Image generation and editing must follow a predefined token order, making localized editing difficult. Practical solutions rely on masking or token reordering, but cannot fully recover bidirectional conditioning.

\paragraph{Hybrid models.}
Hybrid approaches combine AR-based text generation with continuous diffusion or flow-based image synthesis (e.g., BAGEL \citep{pmlr-v235-murty24a}, Transfusion \citep{zhou2025transfusion}). These models decouple semantic reasoning from visual generation, but require trajectory-level modeling and multi-step optimization, making them fundamentally different from both AR and discrete diffusion formulations.

\subsection{Multi-Modal Discrete Diffusion Models}
The generative process with discrete diffusion iteratively predicts the clean state $\mathbf x_0$ over masked positions given $\mathbf x_{t}$ as a input at time $t\in(0, 1]$. 
Starting from the fully masked sequence $\mathbf{x}_T$, a shared Transformer backbone predicts the categorical distribution over clean tokens, $p_\theta(\mathbf{x}_0 \mid \mathbf{x}_t)$, for all masked positions. The total denoising step $K$ defines the sampling schedule with time $t$, and the noised sequence $\mathbf x_t$ is iteratively remasked using confidence score of the model until it reaches the fully denoised state at $\mathbf x_0$.

Multi-modal discrete diffusion models receive both image and text as inputs, and are capable of generating the image and text tokens in discrete space. More specifically, the input images are reconstructed into a sequence of visual tokens with image tokenizer, where MMaDA-Parallel uses MagVIT-v2 \citep{yu2024language} and LaViDa-O uses the encoder of Meissonic \citep{bai2025meissonic}, while text is processed by the discrete diffusion backbones tokenizer like LLaDA-8B \citep{nie2026large}. Both image and text tokens are jointly modeled within a unified probabilistic framework.

\subsection{Visual Thinking and Intermediate Visual Reasoning}

\paragraph{External tool-based methods.}
\citet{zhang2026thyme,zhou2025reinforced,ding2025arm,zheng2026deepeyes,wang2025pixel} augment models with explicit visual operations (e.g., zooming, cropping, region selection). These approaches operate directly in pixel space and iteratively refine visual evidence. They typically rely on reinforcement learning (e.g., GRPO) to encourage tool usage, as supervised fine-tuning alone is insufficient.

\paragraph{Internal latent reasoning.}
Some works \citep{yang2025machine,li2025latent} perform reasoning over internal visual representations without explicit image manipulation. Training typically combines SFT with RL to stabilize multi-modal reasoning trajectories.

\paragraph{Intermediate image generation.}
Other approaches explicitly generate intermediate images as part of reasoning. For example, \citet{chern2025thinking} produces visual subgoals and self-refined hypotheses, while \citet{xu2026visual} formulates reasoning as sequences of generated visual states. These methods often rely on synthetic SFT data combined with RL to encourage coherent multi-step reasoning.

Visual thinking methods emphasize iterative reasoning, bidirectional context, and explicit spatial grounding. However, they typically rely on external tools or multi-step interaction and do not optimize a unified generative likelihood. In contrast, unified AR and DDM models operate within a single generative framework. Notably, the need for bidirectional conditioning and flexible editing highlighted by these works aligns more naturally with diffusion-based approaches than with auto-regressive models.

\subsection{GRPO with Discrete Diffusion Models}
In AR language models, GRPO relies on an exact token-level likelihood factorization, enabling unbiased importance sampling ratios and a sum of clipped policy gradients over tokens. In contrast, discrete diffusion language models generate text via iterative denoising without tractable sequence likelihoods, making GRPO’s importance sampling ratio term approximate \citep{zhaod1,tang2026wd}. Consequently, diffusion variants of GRPO fundamentally shift from token-level, causally factorized optimization to trajectory-level objectives over denoising processes.

For language-only diffusion, d1 \citep{zhaod1} extends GRPO to masked diffusion by introducing a surrogate likelihood-based objective over denoised sequences; wd1 \citep{tang2026wd} replaces unstable importance ratios with a ratio-free weighted log-likelihood formulation to reduce variance; and lateral-thought-style method \citep{huang2026reinforcing} shifts optimization to the diffusion chain itself, assigning credit across intermediate denoising states rather than tokens.

\section{Prompt Templates}\label{app: prompt}
The prompt template for supervised finetuning and GRPO phase is shared to induce the desired behavior from the model at rollout. These are the list of templates used for training:
\begin{itemize}
    \item \textbf{Grounding ($P_\text{ground}$):} \texttt{Your job is to identify the region where auxiliary line, box, or editing could help solve the following problem. Give bounding boxes in LOC format.}
    \item \textbf{Visual Reasoning ($P_\text{visual}$):} \texttt{Edit the region where auxiliary line, box, or drawing could help solve the following problem.}
    \item \textbf{Textual Reasoning ($P_\text{textual}$):} \texttt{Let's think step-by-step to solve the question. Put your final answer in <answer> </answer> tags.}
\end{itemize}

\section{SFT Dataset: \dataset{}}
\label{app: dataset construction}
To localize such edits, we first align each intermediate reasoning image with the original image by estimating a homography transformation using Oriented FAST and rotated BRIEF (ORB) features \citep{10.1109/ICCV.2011.6126544}, establishing pixel-wise spatial correspondence under geometric variations. We then compute a difference map between the aligned images. To improve robustness, we apply Gaussian smoothing to suppress noise and minor pixel-level discrepancies while preserving structured edits. The resulting map is thresholded and post-processed to extract a tight bounding box covering the most salient modified region. This procedure yields region-level supervision aligned with reasoning steps, enabling the model to associate intermediate reasoning with precise spatial edits and improving its readiness for grounded editing in the subsequent RL stage.

The dataset utilizes a normalized spatial representation where raw ground truth coordinates, $\mathbf{B}_\text{raw} = [x_\text{min}, y_\text{min}, x_\text{max}, y_\text{max}]$, are transformed from the original image dimensions to a standardized target resolution $H \times W$. For integration into the LaViDa-O Supervised Fine-Tuning (SFT) framework, these coordinates are serialized into a discrete \texttt{LOC} string format: the sequence is enclosed by structural tokens $\langle\text{LOC\_BEGIN}\rangle$ and $\langle\text{LOC\_END}\rangle$, with each integer coordinate $v_i$ represented as $\langle\text{LOC\_}v_i\rangle$. For instance, a raw box $[8, 41, 794, 69]$ from an $800 \times 557$ image is scaled to $[10, 207, 1016, 243]$ and encoded as: \texttt{<LOC\_BEGIN><LOC\_10><LOC\_207><LOC\_1016><LOC\_243><LOC\_END>}.

\begin{figure}[bh]
    \centering
    \includegraphics[width=0.7\linewidth]{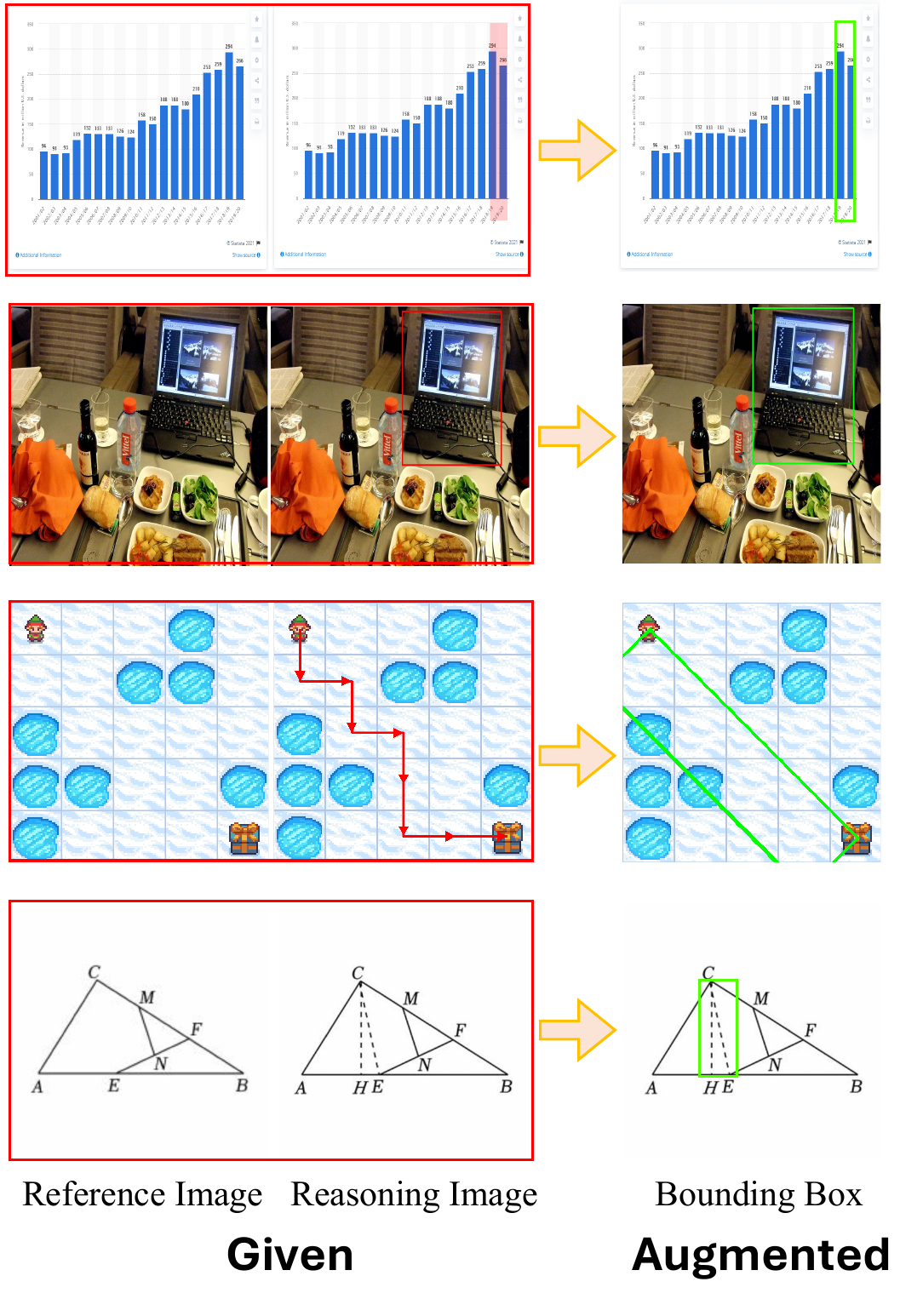}
    \caption{\dataset{} \textbf{Sample.}}
    \label{fig:dataset}
\end{figure}

In \cref{fig:dataset}, we show samples from the augmented dataset, \dataset{}. We construct the dataset by placing bounding boxes over the regions that differ between the ground-truth reasoning image and the original problem image. The dataset includes 6K samples each from ThinkMorph Spatial Navigation, Chart Refocus, and Visual Search, as well as 86.5K samples from Math-VR Train \citep{duan2025codeplot}. In total, \dataset{} contains 104.5K triplets consisting of a reference image, a ground-truth reasoning image, and a bounding box annotation.

Note that Math-VR Train is not used for either SFT or RL training in this paper. However, we will release the dataset to support future research.

\clearpage
\section{Training Details}\label{app: Training Details}

\subsection{Supervised Fine-Tuning Details}
\paragraph{Dataset Composition for SFT.}
From \textit{ZebraCoT}, we randomly sample 1{,}000 instances from a pool of 9{,}000 examples spanning diverse reasoning tasks (e.g., visual puzzles, board games, symbolic reasoning, and planning). Each instance is converted into two variants: (i) text-based reasoning and (ii) interleaved multi-modal reasoning with both image and text inputs. From \textit{ThinkMorph}, we randomly sample 1{,}000 instances from tasks including chart refocus, spatial navigation, visual search, and blink-jigsaw.  Total of 9,000 instances, proliferated into two versions are used for supervised fine-tuning.

\paragraph{SFT Hyperparameters.}
We summarize the SFT hyperparameters in Table~\ref{tab:sft_hyperparams}. The model is trained for 10 epochs with bf16 precision, cosine learning-rate decay with warmup, and a maximum sequence length of 8192 tokens.

\begin{table*}[bh]
    \centering
    \small
    \setlength{\tabcolsep}{8pt}
    \renewcommand{\arraystretch}{1.15}
    \begin{tabular}{@{}lcl@{}}
        \toprule
        \textbf{Hyperparameter}
        & \textbf{Setting}
        & \textbf{Description} \\
        \midrule

        \multicolumn{3}{@{}l}{\textbf{Training Setup}} \\
        \midrule
        Number of GPUs                  & 8 (NVIDIA H100)                         & distributed training setup \\
        Precision                       & bf16                      & mixed-precision training \\
        DeepSpeed Stage                 & ZeRO-2                    & optimizer/state sharding \\
        Number of Epochs                & 10                       & total training epochs \\
        \midrule

        \multicolumn{3}{@{}l}{\textbf{Optimization}} \\
        \midrule
        Learning Rate                   & $5\times10^{-6}$          & peak learning rate \\
        Vision Tower Learning Rate      & $2\times10^{-6}$          & vision encoder learning rate \\
        Weight Decay                    & 0.0                       & L2 regularization \\
        Warmup Ratio                    & 0.03                      & warmup proportion \\
        LR Scheduler                    & cosine    & decay schedule \\
        \bottomrule
    \end{tabular}
    \caption{SFT hyperparameters used in our experiments. Settings are taken from the training configuration.}
    \label{tab:sft_hyperparams}
\end{table*}

\clearpage
\subsection{GRPO Details}
\paragraph{GRPO Hyperparameters.}
The hyperparameter settings for GRPO-based reinforcement learning experiments are reported in Table~\ref{tab:hyperparams}.

\begin{table*}[bh]
    \centering
    \small
    \setlength{\tabcolsep}{8pt}
    \renewcommand{\arraystretch}{1.15}
    \begin{tabular}{@{}lcl@{}}
        \toprule
        \textbf{Hyperparameter} 
        & \textbf{Setting} 
        & \textbf{Description} \\
        \midrule

        \multicolumn{3}{@{}l}{\textbf{Objective}} \\
        \midrule
        KL Loss Coefficient          & 0.04            & KL regularization weight \\
        PPO Clip Range (Low)         & 0.20             & lower clipping bound ($\epsilon$) \\
        PPO Clip Range (High)        & 0.28             & upper clipping bound \\
        \midrule

        \multicolumn{3}{@{}l}{\textbf{Batching}} \\
        \midrule
        Train Batch Size             & 64               & prompts per update \\
        Generations per Prompt       & 8                & sampled responses per prompt \\
        Effective Batch Size         & 512              & $64 \times 8$ rollouts \\
        
        \midrule

        \multicolumn{3}{@{}l}{\textbf{Optimization}} \\
        \midrule
        Learning Rate                & $1\times10^{-5}$ & peak LR \\
        Warmup Ratio                 & $1\times10^{-4}$ & LR warmup proportion \\
        Adam $\beta_1, \beta_2$      & (0.9, 0.99)      & optimizer momentum terms \\
        Weight Decay                 & 0.10             & L2 regularization \\
        Gradient Clipping            & 1.0              & global-norm clip \\
        LR Scheduler                 & constant+warmup  & schedule type \\
        \midrule

        \multicolumn{3}{@{}l}{\textbf{Rollout}} \\
        \midrule
        Temperature                  & 0.6              & sampling temperature \\
        Max Prompt Length            & 2048             & input token limit \\
        Max Completion Length        & 512              & output token limit \\
        Diffusion Steps              & 256              & steps for image generation \\
        Image Edit Steps             & 64               & iterative edit refinement \\
        Image Generation Size           & 1024 (LaViDa-O) / 512 (MMaDA-Parallel \& ANOLE)  & generated image size \\
        Num Mask Samplings               & 2                & Mask sampling per generation \\
        \bottomrule
    \end{tabular}
    \caption{GRPO training hyperparameters used in our experiments.}
    \label{tab:hyperparams}
\end{table*}

\clearpage

\section{Evaluation Details}\label{app: evaluation details}
\begin{table}[bh]
    \centering
    \caption{Evaluation Datasets and Generation Token Lengths in text. Image rollout tokens is fixed to 1024 for both LaViDa-O and MMaDA.}
    \label{tab:eval_token_length}
    \vspace{2mm}
    \begin{tabular}{lcl}
        \toprule
        \textbf{Dataset Name} & \textbf{Text Token Length} & \textbf{Hugging Face Source (URL)} \\
        \midrule
        MM-Vet & 1024 & \href{https://huggingface.co/datasets/whyu/mm-vet}{huggingface.co/datasets/whyu/mm-vet} \\
        ChartQA & 16 & \href{https://huggingface.co/datasets/lmms-lab/ChartQA}{huggingface.co/datasets/lmms-lab/chartqa} \\
        CV-Bench & 1024 & \href{https://huggingface.co/datasets/Dongyh35/CVBench}{huggingface.co/datasets/Dongyh35/CVBench} \\
        V-STaR Bench & 16 & \href{https://huggingface.co/datasets/V-STaR-Bench/V-STaR}{huggingface.co/datasets/V-STaR-Bench/V-STaR} \\
        BLINK & 1024 & \href{https://huggingface.co/datasets/BLINK-Benchmark/BLINK}{huggingface.co/datasets/BLINK-Benchmark/BLINK} \\
        MMMU (val) & 16 & \href{https://huggingface.co/datasets/MMMU/MMMU}{huggingface.co/datasets/MMMU/MMMU} \\
        \bottomrule
    \end{tabular}
\end{table}

\paragraph{Chain-of-Thought (CoT) Prompts.}
For datasets requiring reasoning, we prepend task-specific CoT instructions. 
\begin{itemize}
    \item  MM-Vet and BLINK: We use a generic step-by-step reasoning prompt - \texttt{First please perform reasoning, and think step by step to provide best answer to the following question.}
    \item CV-Bench: Structured outputs with reasoning enclosed in \texttt{<analysis>} tags and final answers in \texttt{<answer>} tags. - You are a helpful assistant. When user asks a question, your response must include two parts: first, reasoning process enclosed in <analysis>...</analysis> tags, then final answer enclosed in <answer>...</answer> tags. Please provide a clear, concise response within <answer> </answer> tags that directly addresses to question.
\end{itemize}
\clearpage

\section{Qualitative Results}\label{app:qualitative}

\subsection{Visual Reasoning Generations.} 
\begin{figure}[bh]
    \centering
    \includegraphics[width=\linewidth]{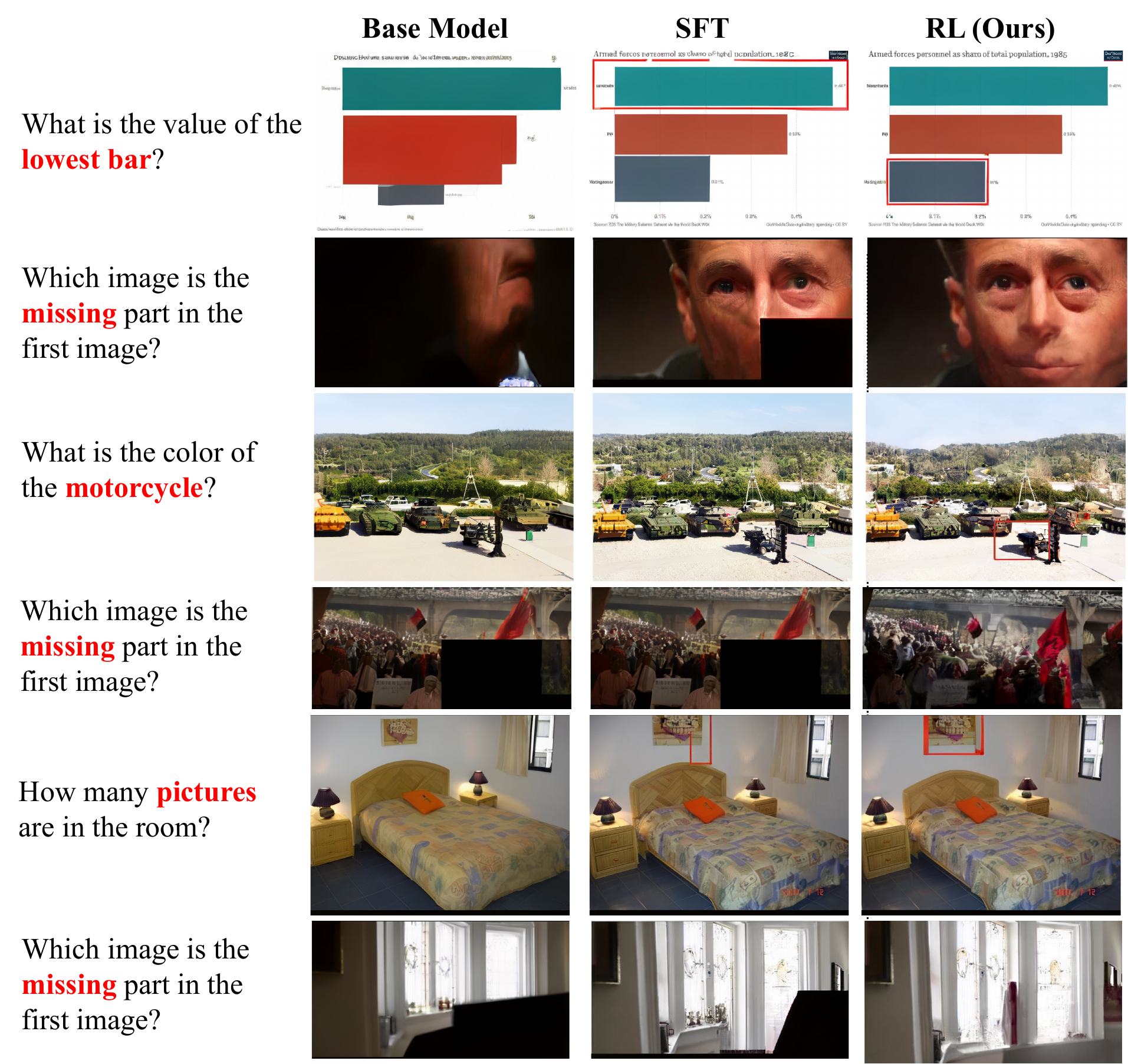}
    \caption{Qualitative visualization of the interleaved reasoning process using localized editing (Part 1). The
model generates intermediate visual proofs while
maintaining visual fidelity. Moreover, the visual fidelity of after RL phase surpasses the SFT finetuned model.}
    \label{fig:qualitative_results}
\end{figure}

\begin{figure}[b]
    \centering
    \includegraphics[width=\linewidth]{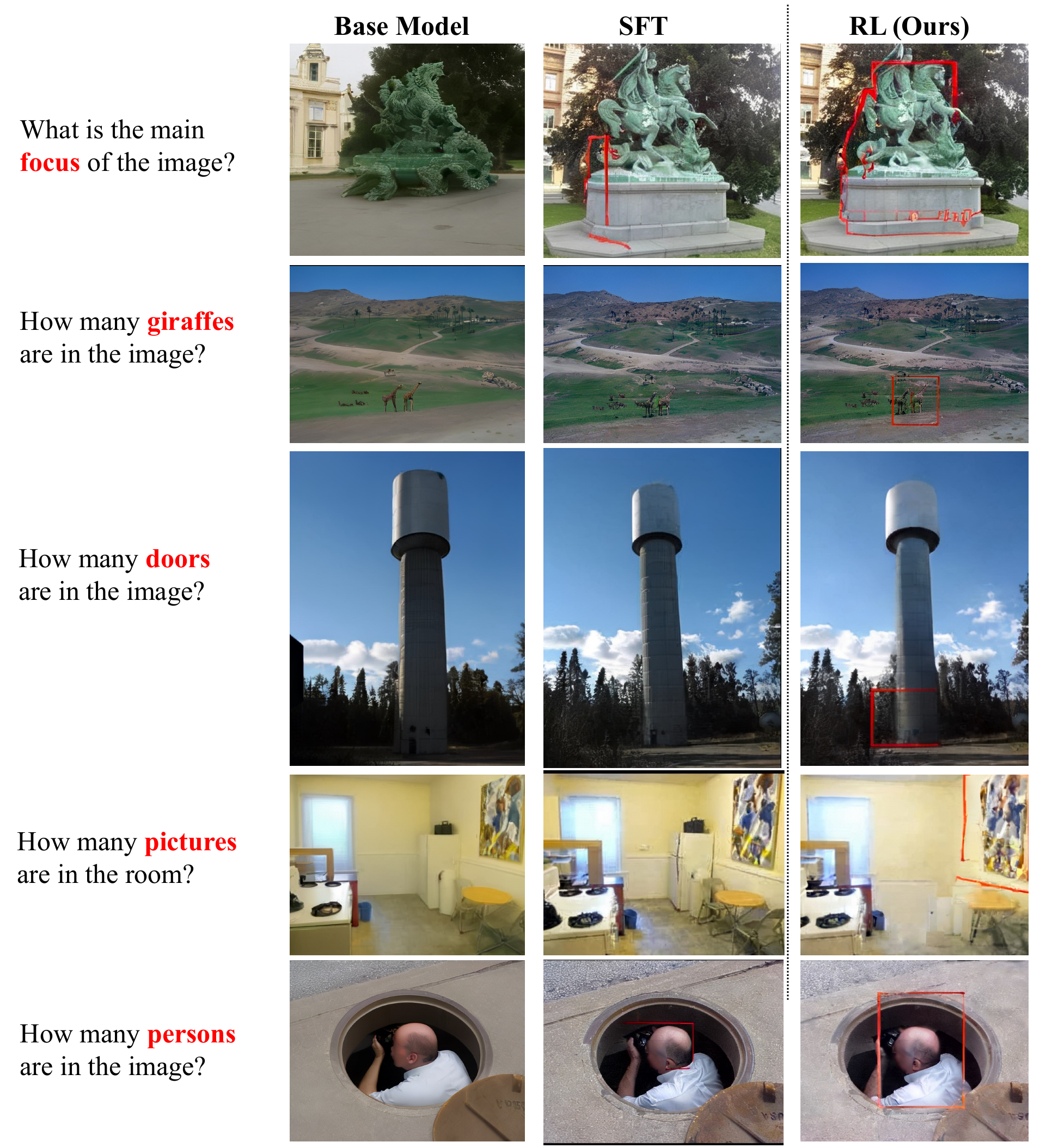}
    \caption{Qualitative visualization of the interleaved reasoning process using localized editing (Part 2).}
    \label{fig:qualitative_results2}
\end{figure}

\begin{figure}[b]
    \centering
    \includegraphics[width=\linewidth]{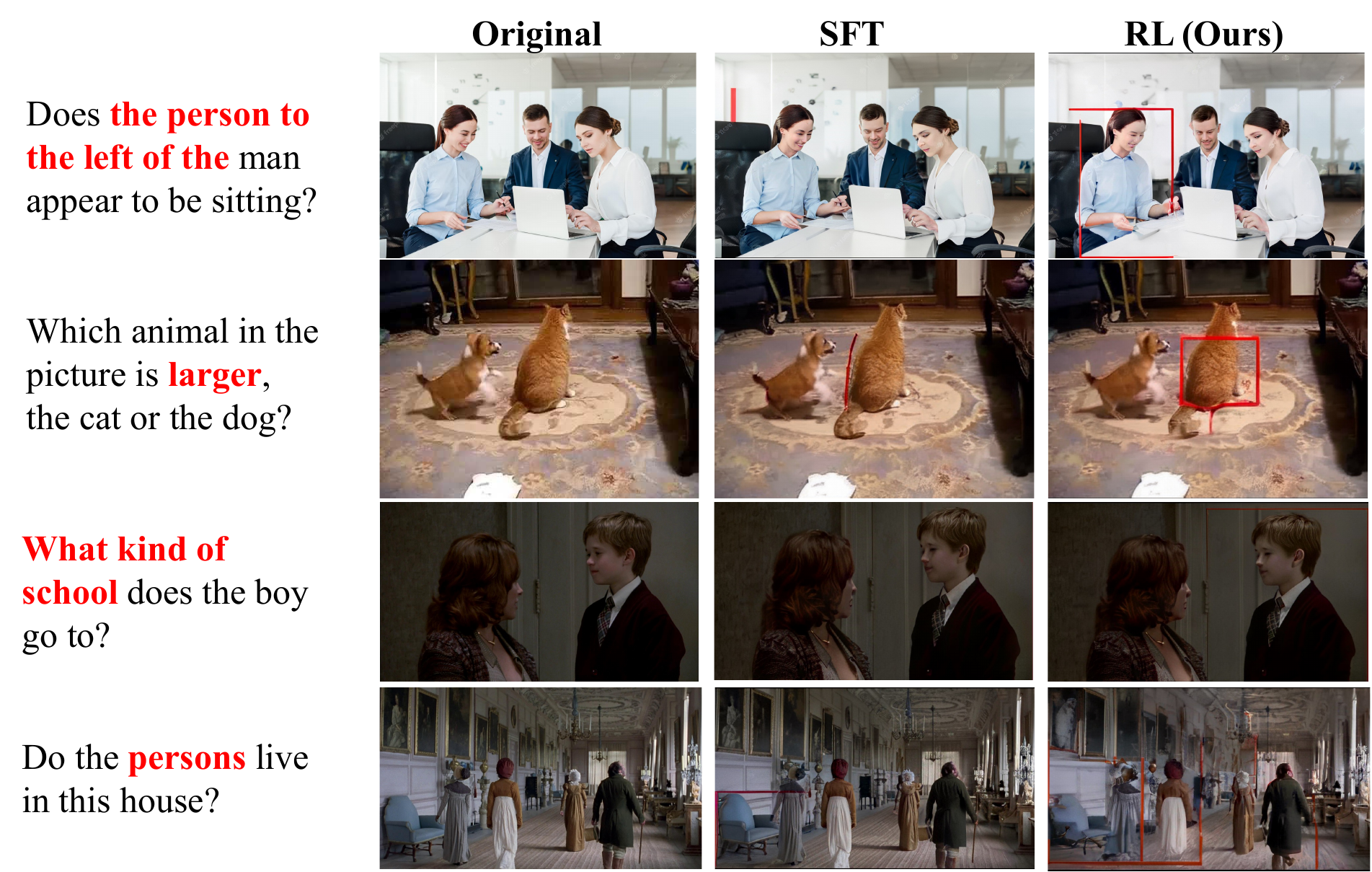}
    \caption{Qualitative visualization of the interleaved reasoning process using localized editing (Part 3).}
    \label{fig:qualitative_results3}
\end{figure}

To evaluate the visual coherence of the proposed method, we provide a qualitative assessment of the intermediate reasoning images generated by the unified models. \Cref{fig:qualitative_results,fig:qualitative_results2,fig:qualitative_results3} illustrates the interleaved generation process, where the model produces localized visual edits to support its linguistic reasoning chain. Despite the significant reduction in computational cost achieved through localized editing, the generated images maintain high semantic alignment with the textual context and visual ground truth. These examples demonstrate that the model successfully preserves fine-grained details and spatial consistency, which are critical for multi-step reasoning tasks that rely on accurate visual feedback during the rollout phase.

\clearpage
\subsection{Failure Analysis}
\begin{figure}[h]
    \centering
    \includegraphics[width=0.7\linewidth]{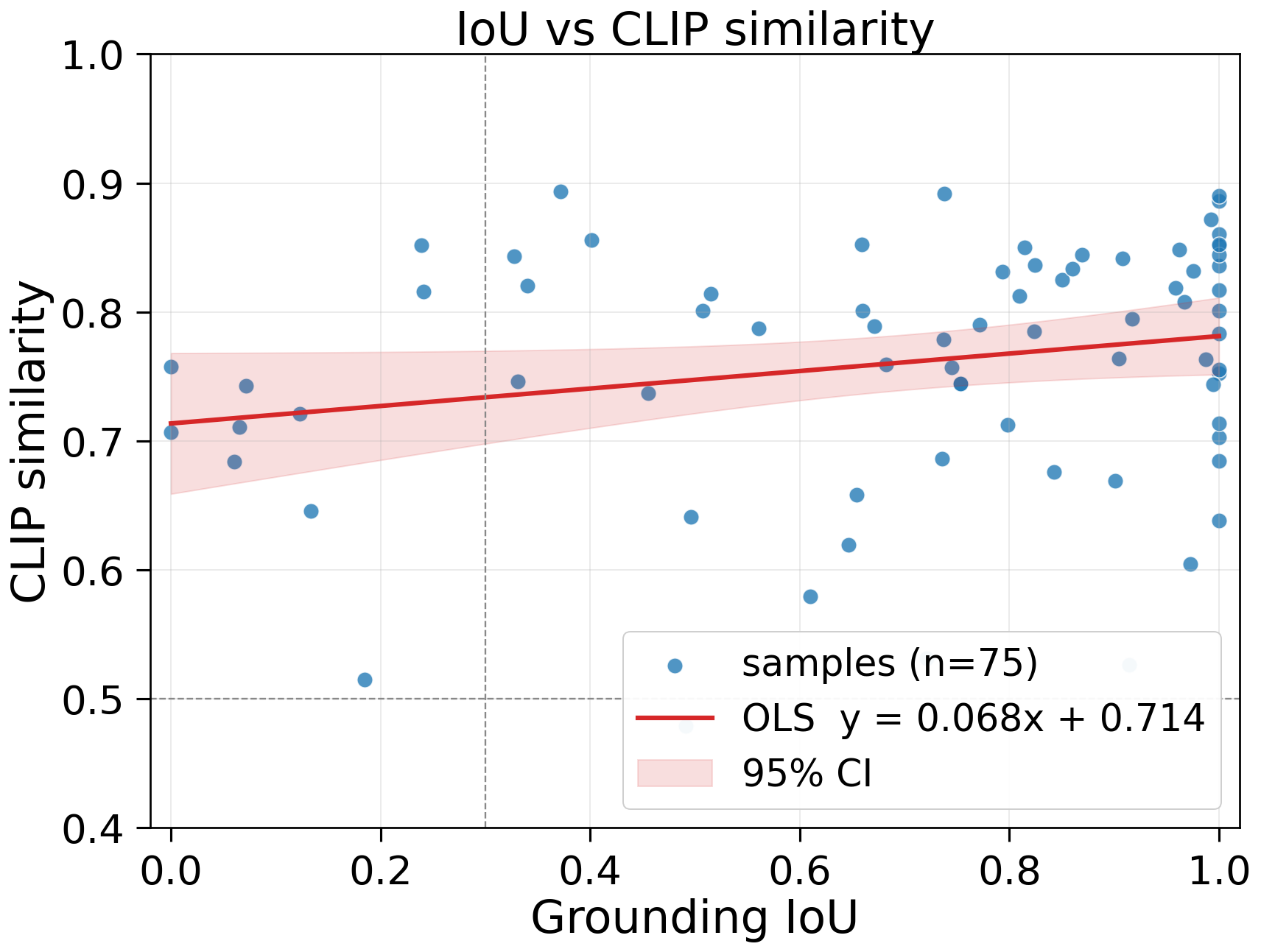}
    \caption{Per-sample relationship between grounding IoU and CLIP image similarity for \ours{}. Each point represents the evaluation samples from ThinkMorph validation set.}
    \label{fig:clip_grnd_scatter}
\end{figure}

In \cref{fig:clip_grnd_scatter}, the regression line between CLIP similarity on the generated visual reasoning and the predicted grounding positions is shows weak positive correlation. Notably, even with completely wrong grounding where IoU is 0, samples are distributed around 0.7 to 0.8 CLIP similarity score. This indicates that for \ours{}, the region the model chose to edit and how visually similar the resulting image is to the ground truth, behave separately.

\begin{figure}[h]
    \centering
    \includegraphics[width=0.65\linewidth]{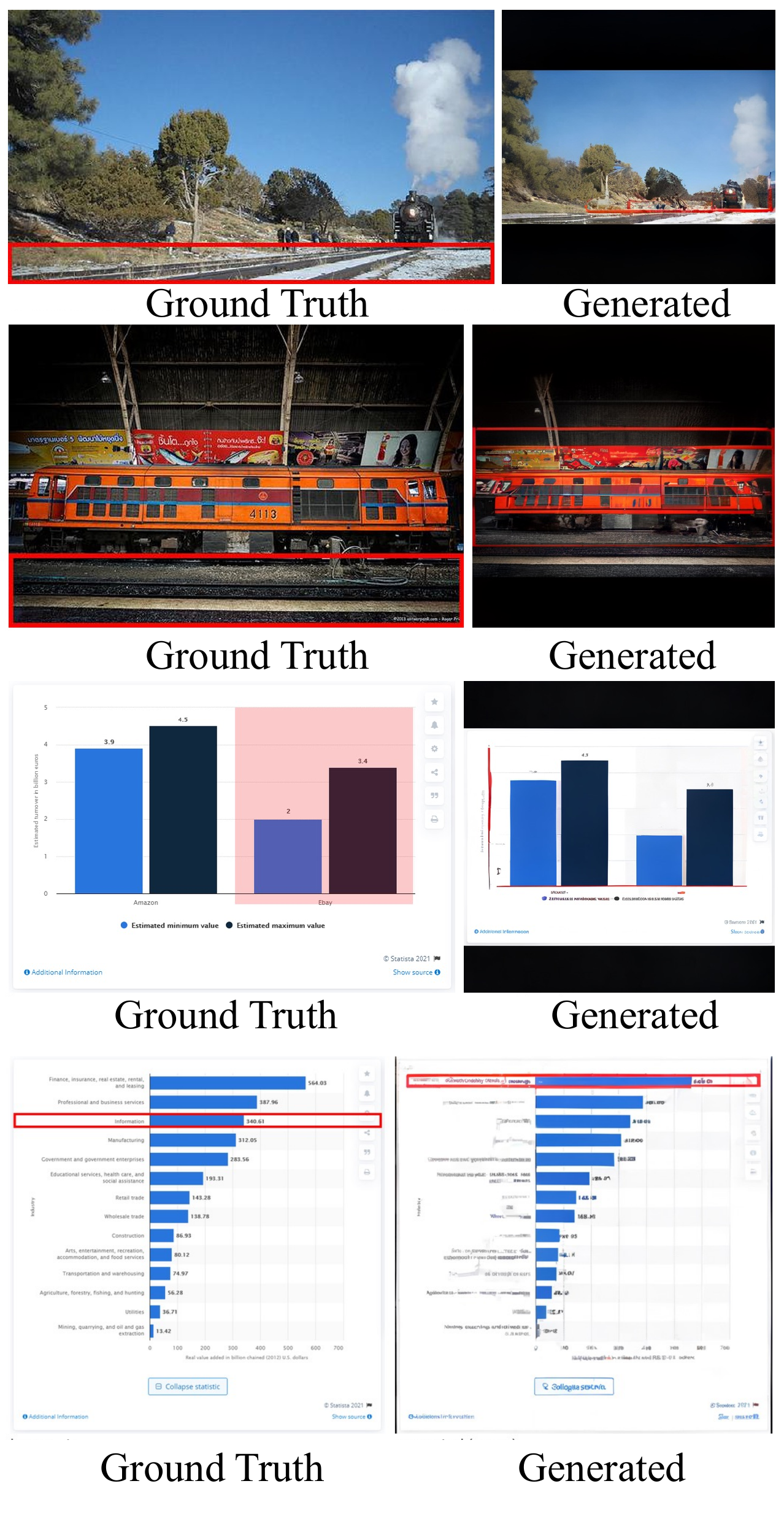}
    \caption{Failure samples with low IoU and CLIP similarity.}
    \label{fig:failures}
\end{figure}

We also demonstrate the worst-4 cases where the IoU score and CLIP similarity are both the lowest in \cref{fig:failures}. The failure in identifying the edit region directly leads to compromised image generation.

\clearpage
\section{Assets and External Resources}

We use several publicly available codebases, pretrained models, and evaluation frameworks in this work. All assets are used in accordance with their respective licenses and intended research usage policies.

\begin{itemize}

    \item \textbf{TRL (Transformer Reinforcement Learning)} from Hugging Face is used for reinforcement learning training and optimization.  
    License: Apache License 2.0.  
    \url{https://github.com/huggingface/trl}

    \item \textbf{d1} is used as a reference implementation for diffusion-based reasoning and GRPO training.  
    License: Apache License 2.0.  
    \url{https://github.com/dllm-reasoning/d1}

    \item \textbf{Anole-7B} is used as a pretrained multimodal foundation model.  
    License: Chameleon License (research-oriented license inherited from Meta Chameleon).  
    \url{https://huggingface.co/GAIR/Anole-7b}

    \item \textbf{MMaDA-Parallel-M} is used as a pretrained multimodal diffusion model.  
    License: MIT License.  
    \url{https://huggingface.co/tyfeld/MMaDA-Parallel-M}

    \item \textbf{LaViDa-O-v1.0} is used as a pretrained multimodal diffusion model.  
    License: Adobe Research License.  
    \url{https://huggingface.co/jacklishufan/LaViDa-O-v1.0}

    \item \textbf{MMaDA-Parallel} inference and implementation codebase.  
    License: MIT License.  
    \url{https://github.com/tyfeld/MMaDA-Parallel/tree/main}

    \item \textbf{LaViDa-O} inference and implementation codebase.  
    License: Adobe Research License.  
    \url{https://github.com/adobe-research/LaVida-O/tree/main}

    \item \textbf{LMMS-Eval} is used as the evaluation framework for multimodal benchmarking.  
    License: Apache License 2.0.  
    \url{https://github.com/evolvinglmms-lab/lmms-eval}

\end{itemize}

We acknowledge and thank the original authors and maintainers of these resources for making their work publicly available. We follow the licenses, usage restrictions, and attribution requirements specified by each respective project and model release.


\newpage
\clearpage
\section*{NeurIPS Paper Checklist}

\begin{enumerate}

\item {\bf Claims}
    \item[] Question: Do the main claims made in the abstract and introduction accurately reflect the paper's contributions and scope?
    \item[] Answer: \answerYes{} 
    \item[] Justification: The contributions and scope in abstract, introduction are properly justified in experiments section (\Cref{sec:experiments}).

\item {\bf Limitations}
    \item[] Question: Does the paper discuss the limitations of the work performed by the authors?
    \item[] Answer: \answerYes{} 
    \item[] Justification: Section 6 has limitations.

\item {\bf Theory assumptions and proofs}
    \item[] Question: For each theoretical result, does the paper provide the full set of assumptions and a complete (and correct) proof?
    \item[] Answer: \answerYes{} 
    \item[] Justification: Section 4.4 has a complete proof and assumptions.

    \item {\bf Experimental result reproducibility}
    \item[] Question: Does the paper fully disclose all the information needed to reproduce the main experimental results of the paper to the extent that it affects the main claims and/or conclusions of the paper (regardless of whether the code and data are provided or not)?
    \item[] Answer: \answerYes{} 
    \item[] Justification: Section 5.1 details the experimental setup and all the hyper-parameter and training variables are disclosed in Appendix.

\item {\bf Open access to data and code}
    \item[] Question: Does the paper provide open access to the data and code, with sufficient instructions to faithfully reproduce the main experimental results, as described in supplemental material?
    \item[] Answer: \answerNo{} 
    \item[] Justification: The authors promise to open-source the code and data in near future.

\item {\bf Experimental setting/details}
    \item[] Question: Does the paper specify all the training and test details (e.g., data splits, hyperparameters, how they were chosen, type of optimizer) necessary to understand the results?
    \item[] Answer: \answerYes{} 
    \item[] Justification: The training and test details are summarized in Appendix B.

\item {\bf Experiment statistical significance}
    \item[] Question: Does the paper report error bars suitably and correctly defined or other appropriate information about the statistical significance of the experiments?
    \item[] Answer:\answerNA{} 
    \item[] Justification: Following the evaluation protocol of previous literature, the evaluation in this paper is conducted with sampling temperature 0 (greedy sampling) that do not require random sampling. Therefore the error bars do not exist for this setting. Also the RL training consumes few days, which is infeasible to conduct multiple trainings, thus rarely required in RL fields.

\item {\bf Experiments compute resources}
    \item[] Question: For each experiment, does the paper provide sufficient information on the computer resources (type of compute workers, memory, time of execution) needed to reproduce the experiments?
    \item[] Answer: \answerYes{} 
    \item[] Justification: In appendix Section D, table 7.
    
\item {\bf Code of ethics}
    \item[] Question: Does the research conducted in the paper conform, in every respect, with the NeurIPS Code of Ethics \url{https://neurips.cc/public/EthicsGuidelines}?
    \item[] Answer: \answerYes{} 
    \item[] Justification: Yes.

\item {\bf Broader impacts}
    \item[] Question: Does the paper discuss both potential positive societal impacts and negative societal impacts of the work performed?
    \item[] Answer: \answerYes{} 
    \item[] Justification: \answerNA{}
    
\item {\bf Safeguards}
    \item[] Question: Does the paper describe safeguards that have been put in place for responsible release of data or models that have a high risk for misuse (e.g., pre-trained language models, image generators, or scraped datasets)?
    \item[] Answer: \answerNA{} 
    \item[] Justification: \answerNA{}

\item {\bf Licenses for existing assets}
    \item[] Question: Are the creators or original owners of assets (e.g., code, data, models), used in the paper, properly credited and are the license and terms of use explicitly mentioned and properly respected?
    \item[] Answer: \answerYes{} 
    \item[] Justification: In appendix Section G.

\item {\bf New assets}
    \item[] Question: Are new assets introduced in the paper well documented and is the documentation provided alongside the assets?
    \item[] Answer: \answerYes{} 
    \item[] Justification: The model training details, license, limitations are discussed.

\item {\bf Crowdsourcing and research with human subjects}
    \item[] Question: For crowdsourcing experiments and research with human subjects, does the paper include the full text of instructions given to participants and screenshots, if applicable, as well as details about compensation (if any)? 
    \item[] Answer: \answerNA{} 
    \item[] Justification: \answerNA{}

\item {\bf Institutional review board (IRB) approvals or equivalent for research with human subjects}
    \item[] Question: Does the paper describe potential risks incurred by study participants, whether such risks were disclosed to the subjects, and whether Institutional Review Board (IRB) approvals (or an equivalent approval/review based on the requirements of your country or institution) were obtained?
    \item[] Answer: \answerNA{} 
    \item[] Justification: \answerNA{}

\item {\bf Declaration of LLM usage}
    \item[] Question: Does the paper describe the usage of LLMs if it is an important, original, or non-standard component of the core methods in this research? Note that if the LLM is used only for writing, editing, or formatting purposes and does \emph{not} impact the core methodology, scientific rigor, or originality of the research, declaration is not required.
    \item[] Answer: \answerNA{} 
    \item[] Justification: \answerNA{}

\end{enumerate}

\end{document}